\documentclass[10pt,twocolumn,letterpaper]{article}

\usepackage{iccv}

\usepackage{times}
\usepackage{epsfig}
\usepackage{graphicx}
\usepackage{amsmath}
\usepackage{amssymb}
\usepackage{siunitx}
\usepackage{multirow}
\usepackage{gensymb}
\usepackage{caption}
\usepackage{colortbl}
\usepackage{subcaption}
\usepackage{textcomp}
\usepackage{siunitx}
\usepackage[dvipsnames]{xcolor} 
\usepackage[section]{placeins} 
\usepackage{threeparttable,booktabs} 
\usepackage{etoolbox}
\appto\TPTnoteSettings{\footnotesize}

\iccvfinalcopy 

\def\iccvPaperID{2364} 

\ificcvfinal\fi
\begin{document}

\title{
A New Technique of Camera Calibration: \\A Geometric Approach Based on 
Principal Lines
}

\author{Jen-Hui Chuang\textsuperscript{1} \qquad
Chih-Hui Ho\textsuperscript{2} \qquad
Ardian Umam\textsuperscript{1} \qquad \\
HsinYi Chen\textsuperscript{3} \qquad
Mu-Tien Lu\textsuperscript{1} \qquad
Jenq-Neng Hwang\textsuperscript{4} \qquad
Tai-An Chen\textsuperscript{3} \vspace{2pt} \\
{\tt\small \{jchuang@cs.,ardianumam.05g@g2.,ryanlu.cs02g@.\}nctu.edu.tw} \qquad {\tt\small chh279@eng.ucsd.edu} \qquad \\  {\tt\small \{JessieChen, tachen\}@itri.org.tw} \qquad {\tt\small hwang@uw.edu} \vspace{2pt}
\\ 
National Chiao Tung University\textsuperscript{1} \qquad 
University of California, San Diego\textsuperscript{2} \qquad \\
Industrial Technology Research Institute\textsuperscript{3} \qquad 
University of Washington\textsuperscript{4}
}

\maketitle

\begin{abstract}
Camera calibration is a crucial prerequisite in many applications of 
computer vision. In this paper, a new, geometry-based camera calibration 
technique is proposed, which resolves two main issues associated with 
the widely used Zhang's method: (i) the lack of guidelines to avoid 
outliers in the computation and (ii) the assumption of fixed camera 
focal length. The proposed approach is based on the closed-form solution 
of principal lines (PLs), with their intersection being the principal 
point while each PL can concisely represent relative 
orientation/position (up to one degree of freedom for both) between a 
special pair of coordinate systems of image plane and calibration 
pattern. With such analytically tractable image features, computations 
associated with the calibration are greatly simplified, while the 
guidelines in (i) can be established intuitively. Experimental results 
for synthetic and real data show that the proposed approach does compare 
favorably with Zhang's method, in terms of correctness, robustness, and 
flexibility, and addresses issues (i) and (ii) satisfactorily.

\end{abstract}

\section{Introduction}\label{sec:intro}
Camera calibration is a crucial step in many 3D vision applications, such as robotic navigation \cite{slam}, depth map estimation \cite{Hirschmuller05}, and 3D reconstruction \cite{3D_hand}. Camera calibration establishes the geometric relation between 3D world coordinate system (\textbf{WCS}) and the 2D image plane of camera by finding extrinsic and intrinsic camera parameters. The extrinsic parameters, which define the translation and orientation of the camera with respect to the world frame, transform 3D \textbf{WCS} into 3D camera coordinate system (\textbf{CCS}). On the other hand, intrinsic parameters, including principal point, focal length and skewness factor, transform 3D \textbf{CCS} to 2D image plane of the camera.

Camera calibration can be roughly classified into two categories: photogrammetric \cite{Mit93,Tsai87} and self-calibration \cite{Maybank1992, Bougnoux98}. The former methods perform calibration based on sufficient measurements of 3D points with known correspondence in the scene and assume some calibration objects/templates are available. However, in a large-scale camera network, it is hard to acquire this kind of measurement or available information for each camera. Therefore, many methods have been proposed to self-calibrate the camera automatically based on certain assumptions on the online camera scenes \cite{Mohedano10,Tang16}.  On the other hand, both methods can also accomplish camera calibration through vanishing points-based methods \cite{Caprile1990,Wang1991} or pure rotation approaches \cite{Wang1991}.

Zhang’s method \cite{Zhang2000} is considered as the most widely used photogrammetric calibration method due to its low cost and flexibility, which only needs to use a printed pattern of checkerboard pasted on a flat surface, and captured by the camera with at least two different orientations. However, two main issues are associated with such an approach: (i) The checkerboard patterns are usually placed randomly and used all together, without a systematic procedure to screen out ill-posed patterns,
and (ii) all intrinsic parameters are assumed to be fixed throughout the pattern capturing process.
For Issue (i), inconsistent or unreasonable calibration results may be generated from different sets of checkerboard patterns for the same camera, as the two (intrinsic and extrinsic) apparently independent sets of parameters are simultaneously calculated via purely algebraic formulations. Moreover, the complexity of such formulations, which are not established for the original intrinsic parameters but for their nonlinear transformations, also greatly decrease the feasibility of the development of more general formulations to resolve Issue (ii) which may occur quite often in practice.

Tan et al. \cite{Tan2017AutomaticCC} partly address issue (i) by first replacing physical checkerboard patterns with virtual ones displayed on a screen to minimize localization error of point features (the corner points) resulting from a blurry image due to hand motion. Then, by conveniently using different sets of virtual patterns in the experiments, appropriate poses of these virtual patterns are suggested: the selected point features should distribute uniformly across the image captured by the camera.  Nonetheless, such a conclusion may not be decisive as different suggestions for appropriate poses of calibration object/pattern also exist \cite{Rojtberg2017,Ricolfe2011}.

\begin{figure}[t]
  \centering
  \includegraphics[width=0.8\linewidth]{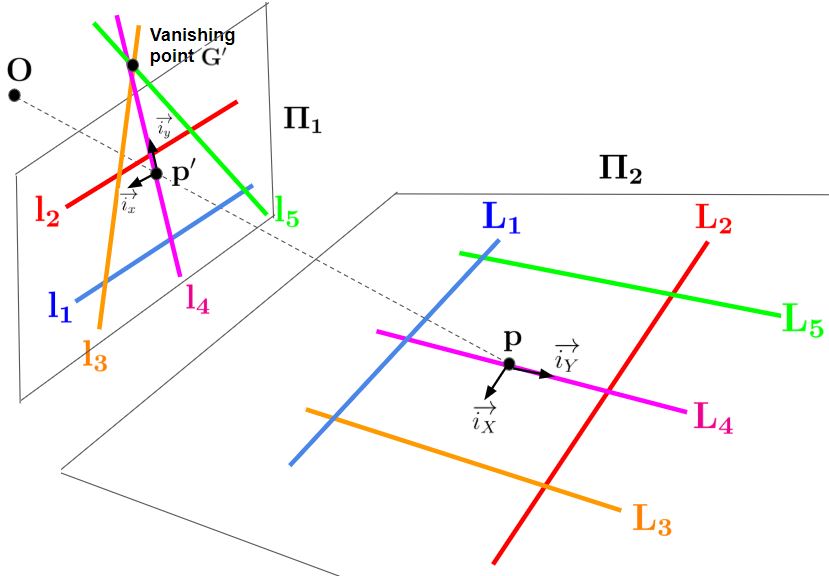}
  \caption{Special geometric relationship of \textbf{WCS} and \textbf{IPCS} for 1D localization of principal point with principal line.}
  \label{fig:wcs_ipcs}
\end{figure}

To partly address Issue (ii), only the principal point is assumed fixed and estimated in \cite{Alturki2016} and \cite{Lu2018}, with the skewness factor ignored and the focal length not assumed to be fixed. The estimation is based on the establishment of a coordinate system of the image plane (\textbf{IPCS}), which has a special geometric relationship to a corresponding \textbf{WCS}, with the X-Y plane (\textit{\textbf{the calibration plane}}) of the latter containing the calibration pattern. Specifically, their relationship can be described by the following rules of geometry, as depicted in Figure \ref{fig:wcs_ipcs}, wherein $\Pi_1$ and $\Pi_2$ are the image plane and the calibration plane, respectively, and the projection center $O$ is colinear with the line connecting the origins of \textbf{IPCS} and \textbf{WCS}:

\noindent \textbf{R1}. $\overrightarrow{i_X}$ is parallel to intersection of $\Pi_1$ and $\Pi_2$.

\noindent \textbf{R2}. Image of $\overrightarrow{i_Y}$(colinear with $\overrightarrow{i_y}$) is perpendicular to $\overrightarrow{i_x}$.


For \textbf{R1}, it is not hard to see that only lines parallel to $\overrightarrow{i_X}$ , e.g., $L_1$ and $L_2$, will have their images parallel to $\overrightarrow{i_x}$. On the other hand, the line containing $\overrightarrow{i_y}$ in \textbf{R2}, e.g., $l_4$ in Figure \ref{fig:wcs_ipcs}, will passes through the vanishing point of images of lines perpendicular to $\overrightarrow{i_X}$, e.g., $L_3$,$L_4$ and $L_5$, and correspond to the axis of symmetry of them. Moreover, it is shown in \cite{Alturki2016} and \cite{Lu2018} that such image line feature, called \textit{\textbf{principal line}} in this paper, will also pass through the principal point $p$, i.e., the intersection of the optical axis and the image plane. Therefore, the camera principal point can be identified as the intersection of principal lines obtained for a set of calibration planes of different poses.

In this paper, a new technique of camera calibration is proposed. The analytically tractable technique is based on \textbf{R1} and \textbf{R2} described in Figure~\ref{fig:wcs_ipcs} and has the following desirable features:

\noindent \textbf{F1. Efficiency} \textemdash While the principal line of each calibration plane is obtained empirically by analyzing a sequence of planar image patterns in \cite{Alturki2016} and \cite{Lu2018} before \textbf{R1} and \textbf{R2} are achieved, equation of the principal line is obtained in closed form in this paper using only a single calibration pattern. 
    
\noindent \textbf{F2. Completeness}
\textemdash By assuming the circular symmetry\footnote{Such assumption is quite reasonable nowadays for a variety of cameras, as one can see later in Sec. \ref{sec:experiment}.} of the imaging system, the proposed approach can derived all intrinsic parameters, while extrinsic ones can be found readily for each calibration plane with respect to the special \textbf{WCS-IPCS} pair shown in Figure~\ref{fig:wcs_ipcs}, but with  their origins located on the camera optical axis, as well as for any \textbf{WCS}-\textbf{IPCS} pair if needed.
    
\noindent \textbf{F3. Robustness/accuracy} \textemdash Based on the geometry associated with each corresponding principal line, effective way of identifying ill-posed calibration planes is developed so that the robustness and accuracy of the calibration can be greatly improved by discarding such outliers, resolving Issue (i) of Zhang’s method \cite{Zhang2000}.
    
\noindent\textbf{F4. Flexibility} \textemdash Without assuming a fixed camera focal length (FL), the proposed approach can find different values of FL adopted in the image capturing process for calibration patterns of different poses, and successfully address Issue (ii) of Zhang’s method.

The rest of this paper will be organized as follows. In the next section, a closed-form solution of the principal line is first derived for the homography matrix,

\begin{equation} \label{eq:homography}
H=
\begin{bmatrix}
    h_{1} & h_{2} & h_{3}  \\
    h_{4} & h_{5} & h_{6}  \\
    h_{7} & h_{8} & h_{9} 
\end{bmatrix},
\end{equation}
which is obtained from corresponding point features on a calibration plane and the image plane. A set of such line features can then be used to determine the principal point and, subsequently, the rest camera parameters. In Section \ref{sec:experiment}, experimental results on both synthetic and real data to demonstrate the superiority of proposed approach in robustness, accuracy, and flexibility, with elaborations of some guidelines for the selection of appropriate poses of calibration planes. Finally, some concluding remarks will be given in Section \ref{sec:conclusion}.

\section{Derivation of Closed-Form Solutions of Camera Parameters} \label{sec:para_derive}
In this section, camera parameters are derived analytically via the establishment of the special geometric relation of \textbf{IPCS} and \textbf{WCS} described in \textbf{F2}. In Section \ref{sec:pp_line}, the derivation of principal line with a single image of the calibration pattern is elaborated, which includes the establishment of $\overrightarrow{i_X}$ and $\overrightarrow{i_x}$ of \textbf{R1} via rotation, followed by finding the principal line (and $\overrightarrow{i_y}$ of \textbf{R2}) using the vanishing point. In Section~\ref{sec:pp_line}, closed-form solutions of intrinsic parameters are derived, which include the derivation of the principal point from a set of principal lines, followed by the derivation of the camera focal length by using the principal point to establish the special \textbf{WCS-IPCS} pair described in \textbf{F2}. Finally, extrinsic parameters of can be obtained easily for such pair of coordinate systems.

\subsection{Deriving Closed-Form Solution of the Principal Line} \label{sec:pp_line}
In this section, in order to simply the derivation of principal line on the image plane, orientation a unit square on the calibration plane is considered. Specifically, the rotation of the square which results in a trapezoidal shape of its image is derived in closed form.  Subsequently, the direction of the two bases of the trapezoid is identified as the direction of $\overrightarrow{i_X}$ and $\overrightarrow{i_x}$ of \textbf{R1}, while the principal line is identified as the line orthogonal to the bases and passing through the intersection of the two legs of the trapezoid.

\subsubsection{Finding the Direction of $\protect\overrightarrow{i_X}$ and $\protect\overrightarrow{i_x}$ of \textbf{R1}}

\begin{figure}[t]
  \centering
  \begin{tabular}{ccc}
    \includegraphics[width=0.25\linewidth]{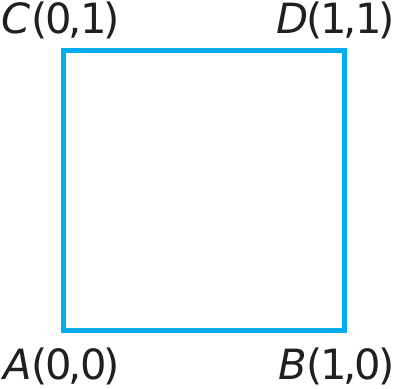} &
    \includegraphics[width=0.3\linewidth]{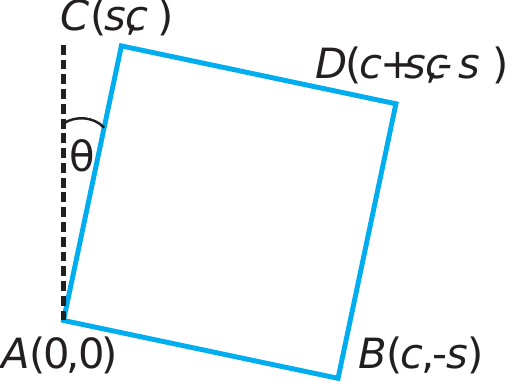} &
    \includegraphics[width=0.25\linewidth]{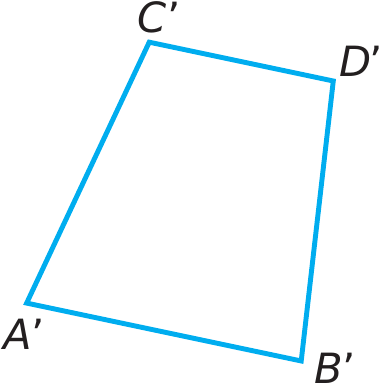} \\
    \footnotesize{(a)}& \footnotesize{(b)}& 
    \footnotesize{(c)}
  \end{tabular}
  \caption{Real world plane projection of a rectangle (a) to the corresponding camera image (c), based on rotation given in (b).
  }
  \label{fig:rot}
\end{figure}

Assume a square $ABCD$ in \textbf{WCS}, as shown in Figure \ref{fig:rot} (a), is captured by a camera, with $A^{\prime}B^{\prime}C^{\prime}D^{\prime}$ being its image in the \textbf{IPCS}. Since $ABCD$ and $A^{\prime}B^{\prime}C^{\prime}D^{\prime}$ are planar surfaces, a homography matrix $\mathbf{H}$ can be used to represent their relationship. The goal of this subsection is to derive the angle $\theta$ in Figure \ref{fig:rot} (b) such that $\overrightarrow{A^{\prime}B^{\prime}}$ is parallel to $\overrightarrow{C^{\prime}D^{\prime}}$, as shown in Figure \ref{fig:rot} (c).

Assume $\mathbf{R}$ is the rotation matrix associated with angle $\theta$, and by rotating rectangle $ABCD$ with $A=\begin{bmatrix}0 & 0 & 1\end{bmatrix}^T$, $B=\begin{bmatrix}1 & 0 & 1\end{bmatrix}^T$, $C=\begin{bmatrix}0 & 1 & 1\end{bmatrix}^T$ and $D=\begin{bmatrix}1 & 1 & 1\end{bmatrix}^T$ using the rotation matrix $R$ with respect to point $A$, i.e.,
\begin{equation}
{\footnotesize R=
\begin{bmatrix}
    \cos(\theta) & \sin(\theta) \\
    -\sin(\theta) & \cos(\theta)
\end{bmatrix} \triangleq
\begin{bmatrix}
    c & s \\
    -s & c
\end{bmatrix}
\nonumber}
\end{equation}
\begin{equation}
{\footnotesize A=
\begin{bmatrix}
0 \\ 0 \\ 1 
\end{bmatrix},
B=
\begin{bmatrix}
c \\ -s \\ 1 
\end{bmatrix},
C=
\begin{bmatrix}
s \\ c \\ 1 
\end{bmatrix},
D=
\begin{bmatrix}
c+s \\ c-s \\ 1 
\end{bmatrix},
\nonumber}
\end{equation}
followed by multiplying with H to transform the rotated rectangle from \textbf{WCS} to \textbf{IPCS}, we have
\begin{equation}
A^{\prime}=
\begin{bmatrix}
\frac{h_3}{h_9} \\ \frac{h_6}{h_9} \\ 1 
\end{bmatrix},
B^{\prime}=
\begin{bmatrix}
\frac{h_1 c-h_2 s+h_3}{h_7 c-h_8 s+h_9} \\ \frac{h_4 c-h_5 s+h_6}{h_7 c-h_8 s+h_9} \\ 1
\end{bmatrix},\nonumber
\end{equation}
\begin{equation}
C^{\prime}=
\begin{bmatrix}
\frac{h_1 s+h_2 c+h_3}{h_7 s+h_8 c+h_9} \\ \frac{h_4 s+h_5 c+h_6}{h_7 s+h_8 c+h_9} \\ 1
\end{bmatrix},
D^{\prime}=
\begin{bmatrix}
\frac{h_1 (c+s)+h_2 (c-s)+h_3}{h_7 (c+s)+h_8 (c-s)+h_9 } \\ \frac{h_4 (c+s)+h_5 (c-s)+h_6}{h_7 (c+s)+h_8 (c-s)+h_9 } \\ 1
\end{bmatrix},\nonumber
\nonumber
\end{equation}
in homogeneous coordinate. By defining $k_1=h_9 (h_7 c-h_8 s+h_9)$ and $k_2=(h_7 c+h_8 s+h_9)(h_7 (c+s)+h_8 (c-s)+h_9)$, $\overrightarrow{A^{\prime}B^{\prime}}$ and $\overrightarrow{C^{\prime}D^{\prime}}$ can be represented as
\begin{equation}\label{eq:ap_bp_IPCS}
\overrightarrow{A^{\prime}B^{\prime}}=
\begin{bmatrix}
\frac{(h_3 h_7-h_1 h_9 )c+(h_2 h_9-h_3 h_8)s}{k_1} \\ \frac{(h_6 h_7-h_4 h_9 )c+(h_5 h_9-h_6 h_8)s}{k_1}
\end{bmatrix}=\begin{bmatrix}
\frac{Uc+Vs}{k_1} \\ \frac{Xc+Ys}{k_1}
\end{bmatrix}
\end{equation}
and
\begin{equation}
\resizebox{.9\hsize}{!}{$
\overrightarrow{C^{\prime}D^{\prime}}=
\begin{bmatrix}
\frac{(h_3 h_7-h_1 h_9 )c+(h_2 h_9-h_3 h_8 )s+(h_2 h_7-h_1 h_8)}{k_2} \\ \frac{(h_6 h_7-h_4 h_9 )c+(h_5 h_9-h_6 h_8)s+(h_5 h_7-h_4 h_8)}{k_2}
\end{bmatrix}=\begin{bmatrix}
\frac{Uc+Vs+W}{k_2} \\ \frac{Xc+Ys+Z}{k_2}.
\end{bmatrix}$},\nonumber
\end{equation}
with
{\footnotesize \begin{align}
U&\triangleq (h_3 h_7-h_1 h_9), 
V\triangleq (h_2 h_9-h_3 h_8),\nonumber\\
W&\triangleq (h_2 h_7-h_1 h_8), 
X\triangleq (h_6 h_7-h_4 h_9),\nonumber\\
Y&\triangleq (h_5 h_9-h_6 h_8), 
Z\triangleq (h_5 h_7-h_4 h_8).\nonumber
\end{align}\par}
Since $\overrightarrow{A^{\prime}B^{\prime}}$ is parallel to $ \overrightarrow{C^{\prime}D^{\prime}}$, we have
\begin{equation}(Uc+Vs)(Xc+Ys+Z) =(Xc+Ys) (Uc+Vs+W)\nonumber\end{equation}
and
\begin{equation} \label{eq:tan_theta}
\tan(\theta)=\frac{s}{c}=\frac{UZ-XW}{YW-VZ}=\frac{h_7}{h_8} 
\end{equation}
Thus, we can obtain the desired rotation angle, 
\begin{equation}\label{eq:rotate_angle}
\theta=tan^{-1}(\frac{s}{c})=tan^{-1}(\frac{h_7}{h_8}). 
\end{equation}

\subsubsection{Finding the Principal Line (and $\protect\overrightarrow{i_y}$ of \textbf{R2})} 
After the rotation shown in Figure~\ref{fig:rot}(b), the intersection between lines $\overrightarrow{AC}$ ($cx-sy=0$) and  $\overrightarrow{BD}$ ($cx-sy-(c^2+s^2)=0$) of the square $ABCD$, denoted by $G$, can be calculated by using the cross product in homogeneous coordinates,
\begin{equation}
G=\begin{bmatrix}
c \\ -s \\ 0
\end{bmatrix} \times \begin{bmatrix}
c \\ -s \\ -(c^2+s^2)
\end{bmatrix} = 
\begin{bmatrix}
s \\ c \\ 0
\end{bmatrix}
\end{equation}
Then, the intersection point in \textbf{IPCS}, denoted by $G^\prime$, which is the vanishing point in \textbf{WCS}, can be obtained by
\begin{equation}\label{eq:g_prime}
G^\prime=HG=\begin{bmatrix}
    \frac{h_1 s+h_2 c}{h_7 s+h_8 c} & \frac{h_4 s+h_5}{h_7 s+h_8 c} & 1
\end{bmatrix}^T.
\end{equation} 
The principal line can be calculated by finding a line that is perpendicular to $\overrightarrow{A^{\prime}B^{\prime}}$ and passing through the vanishing point $G^{\prime}$. By substituting $k_1=1$\footnote{This is obtained by substituting (\ref{eq:rotate_angle}) to $k_1$} and $h_9=1$, (\ref{eq:ap_bp_IPCS}) becomes 
\begin{equation}\label{eq:ap_bp_h9_1}
\overrightarrow{A^{\prime}B^{\prime}}=
\begin{bmatrix}
(h_3 h_7-h_1 )c+(h_2-h_3 h_8)s \\ (h_6 h_7-h_4 )c+(h_5-h_6 h_8)s
\end{bmatrix}.
\end{equation}
Followed by substituting (\ref{eq:rotate_angle}) to (\ref{eq:ap_bp_h9_1}), we have
\begin{align}
\overrightarrow{A^{\prime}B^{\prime}} &=
\begin{bmatrix} \nonumber
-h_1 c+h_2 s \\ -h_4 c+h_5 s
\end{bmatrix} 
= \begin{bmatrix}
-h_1 h_8+h_2 h_7 \\ -h_4 h_8+h_5 h_7
\end{bmatrix}.
\end{align}
A line that is perpendicular to line $\overrightarrow{A^{\prime}B^{\prime}}$ can then be expressed by
\begin{equation} \label{eq:pp_line}
a'u'+b'v'+c'=0,
\end{equation}
where $a'=(-h_1 h_8+h_2 h_7)$, $b'=(-h_4 h_8+h_5 h_7)$ and $c'$ is a constant. The value of $c'$ can be solved by plugging (\ref{eq:g_prime}) to (\ref{eq:pp_line}), so (\ref{eq:pp_line}) not only perpendicular to $\overrightarrow{A^{\prime}B^{\prime}}$ but also passing though vanishing point $G^{\prime}$. The algebraic solution of $c$ is listed in the supplementary material.

Thus, a principal line is derived using \textit{\textbf{only}} homography matrix alone. By repeating the abovementioned procedure for all image set containing calibration pattern, multiple principal lines can be obtained and used to find the principal point, as discussed next.

\subsection{Deriving Closed-Form Solution of Intrinsic Parameters} \label{sec:intrinsic_para}

\subsubsection{Derivation of the Principal Point} \label{sec:pp_point}
Given $n$ principal lines from section~\ref{sec:pp_line} from different calibration poses, with each lines being represented by $a'_i u'+b'_i v'+c'_i=0$ (\ref{eq:pp_line}), the principal point ($u'_0$,$v'_0$) in \textbf{IPCS} can now be estimated as their intersection via the least squares solution 
, or
\begin{equation}\label{eq:pp_sol}
\begin{bmatrix}
u'_0 \\ v'_0
\end{bmatrix}=(D^\mathsf{T}D)^{-1}D^\mathsf{T}C,
\end{equation}
where
\begin{equation}
\resizebox{.9\hsize}{!}{$
D=
\begin{bmatrix}
a'_1 & a'_2 & \hdots & a'_n\\
b'_1 & b'_2 & \hdots & b'_n
\end{bmatrix}^T \nonumber,
C=
\begin{bmatrix}
-c'_1 & -c'_2 & \hdots & -c'_n 
\end{bmatrix}^T$}. \nonumber
\end{equation}

\subsubsection{Derivation of the Focal Length}\label{sec:focal_length}
Considering the original \textbf{IPCS} and \textbf{WCS} coordinate systems, the relationship between point $p$ on the calibration plane ($Z=0$) and its image $p^{\prime}$ can be expressed as

{\footnotesize \begin{align} \label{eq:wcs_ipcs}
p^{\prime}=& M_{int}M_{ext} p= M_{int}\begin{bmatrix}\mathbf{R} & \mathbf{T} \\ \mathbf{0} & 1
\end{bmatrix}p \nonumber\\
=& \begin{bmatrix} f & 0 & u'_0 & 0 \\ 0 & f & v'_0 & 0 \\ 0 & 0 & 1 & 0 \end{bmatrix}\begin{bmatrix} r_{11} & r_{12} & r_{13} & t_X \\ r_{21} & r_{22} & r_{23} & t_Y \\ r_{31} & r_{32} & r_{33} & t_Z \\ 0 & 0 & 0 & 1 \end{bmatrix}\begin{bmatrix}u\\v\\0\\1\end{bmatrix}, 
\end{align}
or with 2D coordinates
\begin{align}\label{eq:wcs_ipcs_2d}
p^{\prime}=& \begin{bmatrix} f & 0 & u'_0 \\ 0 & f & v'_0 \\ 0 & 0 & 1\end{bmatrix}\begin{bmatrix} r_{11} & r_{12} & t_X \\ r_{21} & r_{22} & t_Y \\ r_{31} & r_{32} & t_Z \end{bmatrix} \begin{bmatrix}u\\v\\1\end{bmatrix} \nonumber\\
\triangleq& \tilde{M}_{int}\tilde{M}_{ext} p \nonumber\\
=& \begin{bmatrix} fr_{11}+u'_0 r_{31} & fr_{21}+u'_0r_{32} & ft_X+u'_0t_Z \\ fr_{21}+u'_0 r_{31} & fr_{22}+v'_0r_{32} & ft_Y+v'_0t_Z \\ r_{31} & r_{32} & t_Z \end{bmatrix}p \nonumber\\
=& Hp, 
\end{align}}
where $H$ is the $3\times3$ homography matrix in (\ref{eq:homography}). 

In this section, the formulation of focal length estimation is greatly simplified by transforming \textbf{IPCS} and \textbf{WCS} into the special geometry depicted in Figure~\ref{fig:wcs_ipcs}, with the optical axis passing through their origins.
Note that principal point  $p'_0=\begin{bmatrix}u'_0,v'_0,1\end{bmatrix}^\mathrm{T}$ in \textbf{IPCS} is derived in Section~\ref{sec:pp_point}, whereas its corresponding point $p_0=\begin{bmatrix}u_0,v_0,1\end{bmatrix}^\mathrm{T}$ in \textbf{WCS} can be obtained using (\ref{eq:wcs_ipcs}).
Thus, the coordinate transformation is performed by $shifting$ these origins accordingly, followed by $rotating$ both \textbf{IPCS} and \textbf{WCS} axes according to the principal line (\ref{eq:pp_line}) and its counterpart in \textbf{WCS}, respectively. Therefore, the following formulation can be established from (\ref{eq:wcs_ipcs}):
\begin{align}
    p'_{sr}=&H_{1r}H_{1s}Hp \nonumber\\
    =&H_{1r}H_{1s}H(H_{2r}H_{2s})^{-1}(H_{2r}H_{2s})p \nonumber\\
    =&H_{1r}H_{1s}H(H_{2r}H_{2s})^{-1}p_{sr}\nonumber\\
    \triangleq& H^{new}p_{sr}
\label{eq:shifting_rotation},
\end{align}
where $p'_{sr}$ and $p_{sr}$ are
points of the transformed new \textbf{IPCS} and \textbf{WCS}, respectively, with
\begin{align*}
&H_{1r}=\begin{bmatrix}a' & b' & 0\\ -b' & a' & 0\\0 & 0 & 1\end{bmatrix},\, H_{1s}=\begin{bmatrix}1 & 0 & -u'_0\\0 & 1 & -v'_0\\ 0 & 0 & 1\end{bmatrix}
\end{align*}
and $H_{2r}$ and $H_{2s}$ are similar to $H_{1r}$ and $H_{1s}$, but with $a'$, $b'$, $u'_0$ and $v'_0$ replaced by $a$, $b$, $u_0$ and $v_0$ respectively. Note that $\begin{bmatrix}a' & b'\end{bmatrix}$ and $\begin{bmatrix} a & b\end{bmatrix}$ are coefficients associated with (\ref{eq:pp_line}) and its counterpart in \textbf{WCS}, respectively, with $\begin{bmatrix}a & b & c\end{bmatrix}^T=H^T \begin{bmatrix}a' & b' & c'\end{bmatrix}^T$.
On the other hand, for the new \textbf{IPCS} and \textbf{WCS}, it is easy to see that we will have rotation matrix
{\footnotesize \begin{equation} \label{eq:rot_zero}
\mathbf{R^{new}} = R_Z(\alpha)R_Y(\beta)R_X(\gamma)
=\begin{bmatrix}1 & 0 & 0\\ 0 & cos(\gamma) & -sin(\gamma)\\ 0 & sin(\gamma) & cos(\gamma)\end{bmatrix},
\end{equation}}
or $\alpha=\beta=0$, for their relative orientation, and
\begin{align}
    \label{eq:trans_zero}
    \mathbf{T^{new}}=\begin{bmatrix}0 &0 & t_z^{new}\end{bmatrix} 
\end{align}
for their relative location.
By comparing (\ref{eq:rot_zero}) and (\ref{eq:trans_zero})
with (\ref{eq:wcs_ipcs_2d}),
we have, up to a scaling factor $s$\footnote{Equivalently, but less directly, homography matrix similar to that in (\ref{eq:h_new_simplified}) can be obtained by finding coordinates of point features for the new \textbf{IPCS-WCS} before such matrix can be estimated.},
\begin{equation}\label{eq:h_new_simplified}
    \resizebox{.87\hsize}{!}{$
    H^{new} = \begin{bmatrix}h^{new}_{11} & h^{new}_{12} & h^{new}_{13}\\h^{new}_{21} & h^{new}_{22} & h^{new}_{23}\\h^{new}_{31} & h^{new}_{32} & h^{new}_{33}
    \end{bmatrix}=s\begin{bmatrix}
    f & 0 & 0 \\
    0 & fcos(\gamma) & 0 \\
    0 & sin(\gamma) & t_Z^{new}
    \end{bmatrix}$},
\end{equation} 

and
\begin{equation}\label{eq:gamma_prime}
    \gamma=\cos^{-1}(\frac{h^{new}_{22}}{h^{new}_{11}})
\end{equation}
\begin{equation}\label{eq:tz}
    t_z^{new}=\frac{h^{new}_{33}}{s}
\end{equation}
\begin{equation}\label{eq:focal_sol}
    f=\frac{h^{new}_{11}}{s}
\end{equation}
with $s=\frac{h^{new}_{32}}{sin(\gamma)}.$

\subsection{Derivation of Extrinsic Parameters}
In the previous subsection, analytic expressions of two extrinsic parameters are derived in (\ref{eq:gamma_prime}) and (\ref{eq:tz}). As mentioned before, we have $t_x  = t_y = \alpha = \beta = 0$ for the other four parameters for the new \textbf{WCS-IPCS} pair 
. In fact, only five parameters (\ref{eq:h_new_simplified}) are needed to completely specify the relative position and orientation of the two planes. In particular, the relative position of the new \textbf{WCS-IPCS}, can be represented by the distance ($t_z^{new}$) between the two origins, with $t_x = t_y = 0$, while their relative orientation can be represented by: (a) the azimuth angle of the principal line, i.e., 
$tan^{-1}(\frac{b'}{a'})$ 
and (b) the elevation angle ($\gamma$) between the two planes. Such concise formulation of extrinsic parameters is more intuitive and useful. For example, (b) can be used to screen out calibration patterns of bad poses, while a set of good but redundant patterns can be identified using (a), as will be illustrated in some experimental results presented in the next section.

On the other hand, the set of extrinsic parameters similar to the found in Zhang's method for the original \textbf{IPCS} and \textbf{WCS} can also be obtained if needed.
Following the notation
in (\ref{eq:wcs_ipcs_2d}) and plugging the solution of principal point (\ref{eq:pp_sol}) and focal length (\ref{eq:focal_sol}) to $\tilde{M}_{int}$,  $\tilde{M}_{ext}$ can be solved as
\begin{equation}
    \label{eq:external_params}
    \tilde{M}_{ext} = \tilde{M}^{-1}_{int}\frac{H}{s}.
\end{equation}
However, there are minor difference between $\tilde{M}_{ext}$ and $M_{ext}$ as $r_{13}$, $r_{23}$ and $r_{33}$ are not defined in $\tilde{M}_{ext}$ (see $\tilde{M}_{ext}$ in (\ref{eq:wcs_ipcs})). To solve $M_{ext}$ in (\ref{eq:wcs_ipcs}), we first rewrite rotation matrix $\mathbf{R}=\begin{bmatrix}
\mathbf{r_1} & \mathbf{r_2} & \mathbf{r_3}
\end{bmatrix}$, with $\mathbf{r_i}$ being a column vector. By using the property of rotation matrix $\mathbf{R}$, $\mathbf{r_3}$ can be calculated as $\mathbf{r_3}=\mathbf{r_1}\times \mathbf{r_2}$. Thus, the extrinsic parameters $M_{ext}$ for the original \textbf{IPCS} and \textbf{WCS} pair can be solved.

\section{Experimental Results and Discussions}\label{sec:experiment}
In this section, two sets of experimental results will be provided. First, synthetic data are used to: (i) validate the correctness of analytic expressions derived in Section~\ref{sec:para_derive}, and (ii) compare the accuracy of estimated parameters result with Zhang's, with and without noises added to point features in the image plane, as ground truth (GT) are available. Then, calibration results for real data 
are provided for more comprehensive demonstration of the proposed method.
Possibilities of improving the calibration results by screening out inappropriate calibration patterns are also provided for both cases.

\subsection{Performance Evaluation Using Synthetic Data }
\label{sec:3_1}
In this sub-section,
correctness of analytic expressions derived in Sec. 2.2.1 for the principal point 
will first be verified using synthetic data described in the following. It is assumed for simplicity that the camera optical axis is passing through the origin of \textbf{WCS} whose \textit{X}-\textit{Y} plane, e.g., the calibration plane shown in Figure~\ref{fig:wcs_ipcs}, has a fixed rotation angle $\gamma$ with respect to $\overrightarrow{i_X}$, with additional calibration planes obtained by rotating the plane, each time by $\Delta \alpha$, with respect to $\overrightarrow{i_Z}$. Figure~\ref{fig:8_planes_synthetic} shows a set of images obtained for $\gamma=\Delta\alpha=\ang{45}$ with the principal point located exactly at the image center, wherein four corners of a square calibration pattern are used to derive elements of $\mathbf{H}$ in (\ref{eq:homography}) for each (virtual) 640$\times$480 image.

\begin{table}
\centering
  \begin{tabular}{cc}
  \multicolumn{2}{c}{\includegraphics[width=0.90\linewidth]{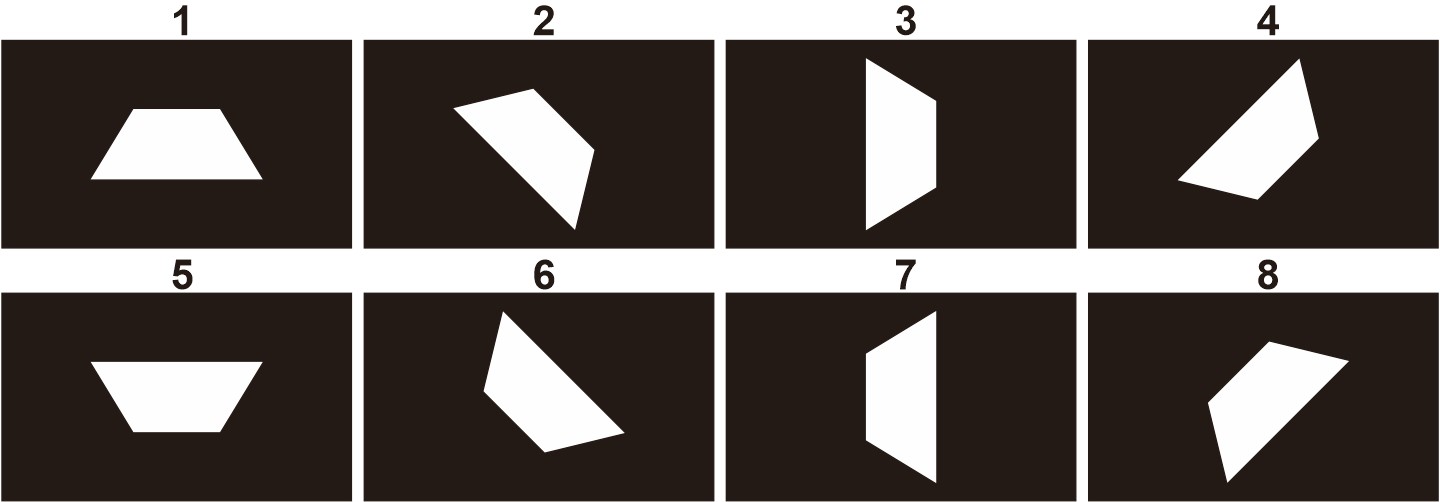}} \\
    \multicolumn{2}{c}{\scriptsize(a)}  \\
    \includegraphics[width=0.4\linewidth]{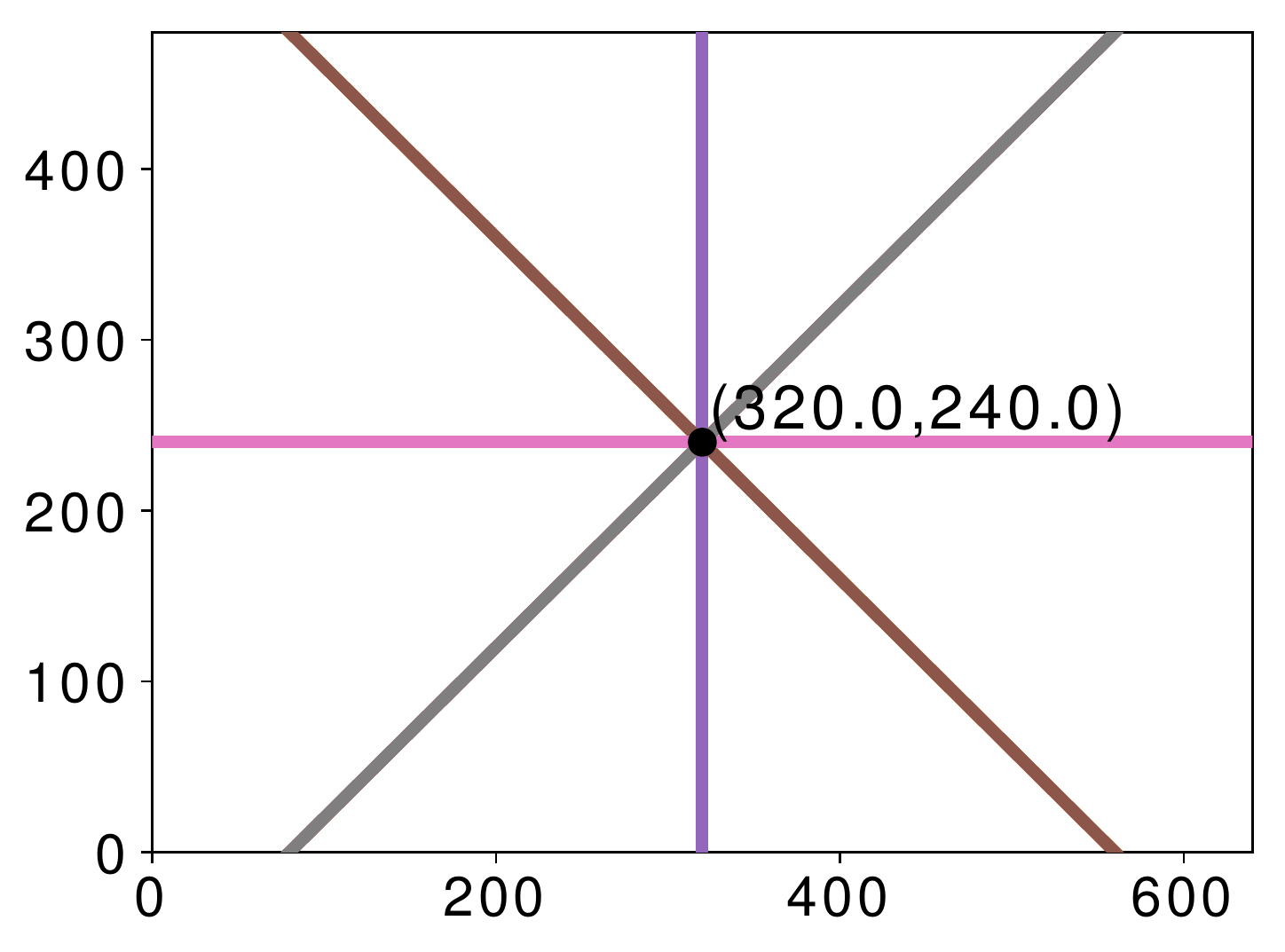} &
    \includegraphics[width=0.4\linewidth]{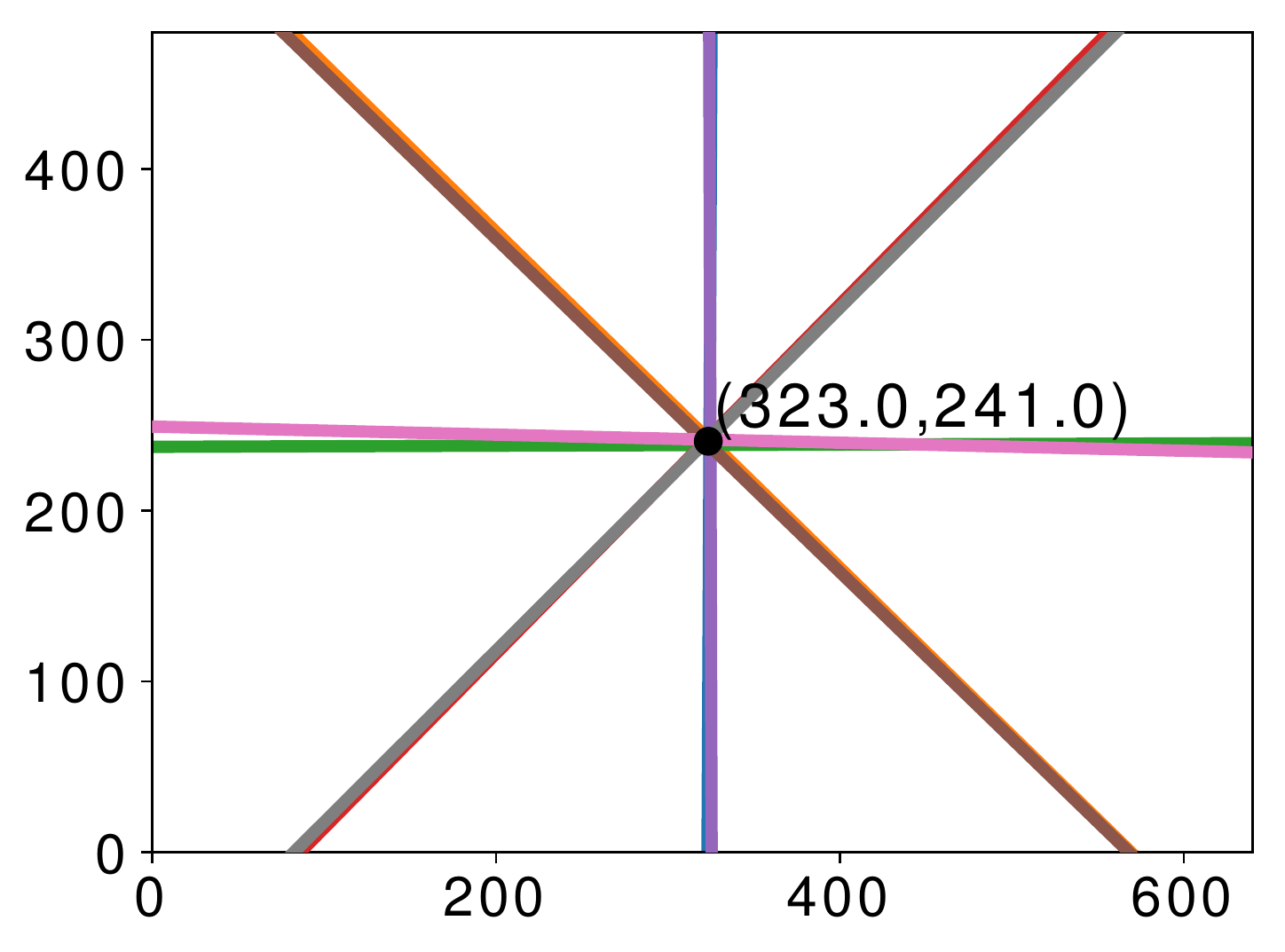}\\
    \scriptsize(b) & \scriptsize(c)
  \end{tabular}
  \captionof{figure}{(a) Eight images obtained for $\gamma=\Delta\alpha=\ang{45}$ (see text). (b) Eight (but merged into four) principal lines. (c) Principal lines obtained with noises.
  }
  \label{fig:8_planes_synthetic}
\end{table}

Multiple sets of images under different extrinsic parameters ($\gamma$, $\beta$, $t_x$, $t_y$ and $t_z$), perturbation (with/without noise) and different focal lengths (FL) are then collected for evaluating the performance of the proposed method and Zhang's method. The evaluation metrics for $PP$, $FL$, rotation $R$~\cite{woit2016quantum} and translation $T$ are based on the difference between the estimation and the ground truth, and can be defined as
\begin{equation}
    \Delta PP = ||PP_{gt}-PP_{est}||_2,
\end{equation}
\begin{equation}
    \label{eq:delta_FL}
    \Delta FL = |FL_{gt}-FL_{est}|,
\end{equation}
\begin{equation}
    \Delta R=cos^{-1}(\frac{\mathbf{Tr}(R_{gt}R_{est}^T)-1}{2}),
\end{equation}
\begin{equation}
    \Delta T = ||T_{gt}-T_{est}||_2.
\end{equation}

\begin{table*}[!htb]
\begin{minipage}{0.44\linewidth}
    \centering
  \begin{tabular}{cc}
    \includegraphics[width=0.44\linewidth]{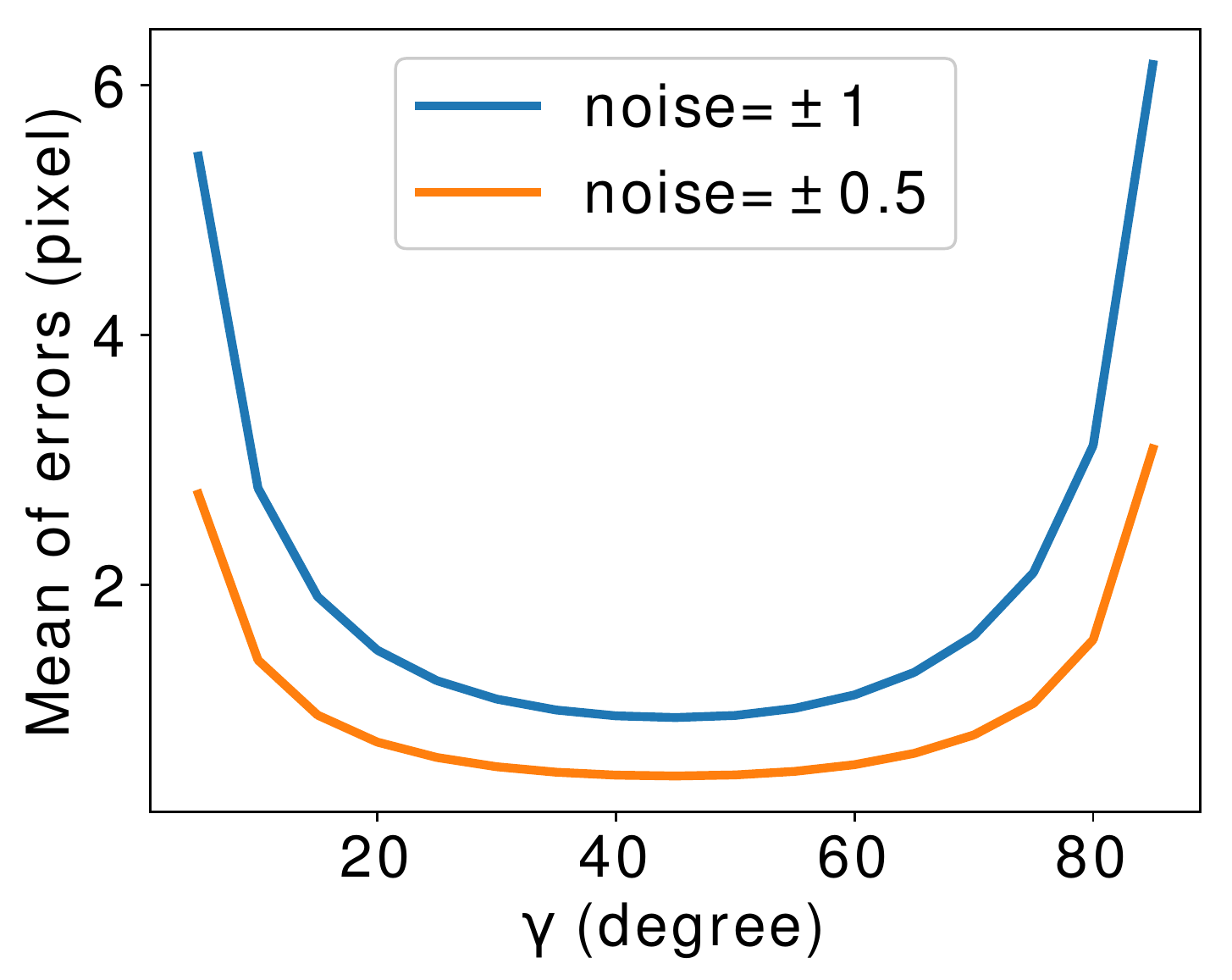} & 
    \includegraphics[width=0.45\linewidth]{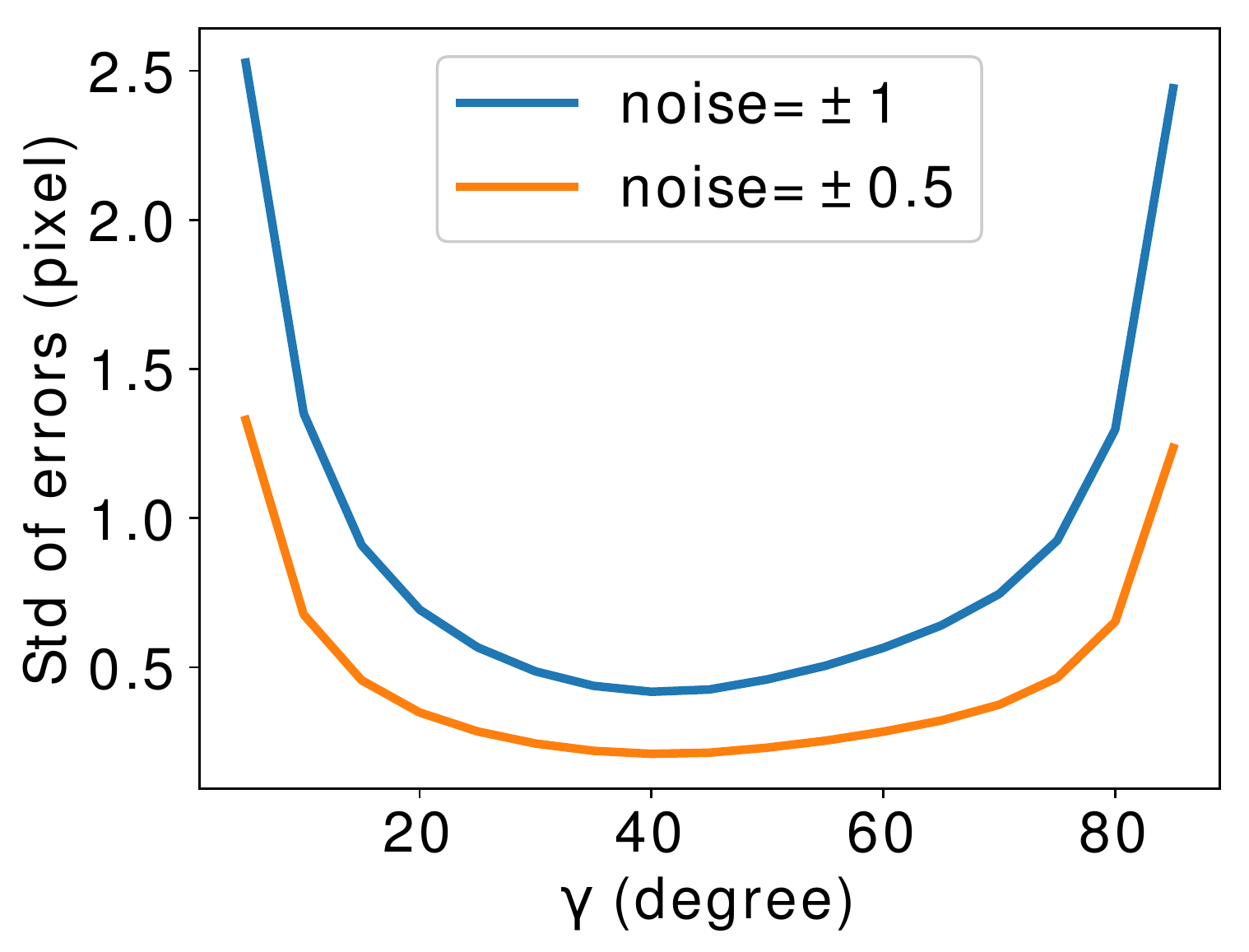}
  \end{tabular}
  \captionof{figure}{(a) Mean values and (b) standard deviations of estimation error for locating the principal point 20 times with two different noise levels, and for different poses of calibration plane.
  }
  \label{fig:pp_err}
  \vspace{-5pt}
\end{minipage}\hspace{10pt}%
\begin{minipage}{0.55\linewidth}
\small \addtolength{\tabcolsep}{-4.4pt}
\centering
\captionof{table}{Comparison of full camera calibration (fixed focal length)}
\label{tab:synthetic_data_ours_vs_zhang}
\begin{threeparttable}
\setlength{\tabcolsep}{0.85pt}
{\renewcommand{\arraystretch}{1}
\begin{tabular}{|c|c|c|c||c|c|c|c||c|c|c|c|}
\hline
\multirow{3}{*}{Set} & \multicolumn{3}{c|}{\multirow{2}{*}{Given params.\tnote{*}}}     & \multicolumn{8}{c|}{Estimation errors}\\ \cline{5-12} 
 & \multicolumn{3}{c|}{}  & \multicolumn{4}{c|}{Ours} & \multicolumn{4}{c|}{Zhang's} \\ \cline{2-12} 
 & $\gamma$, $\beta$ & FL & $t_x$, $t_y$, $t_z$ & $\Delta PP$ & $\Delta FL$ & $\Delta R$ & $\Delta T$ & $\Delta PP$ & $\Delta FL$ & $\Delta R$ & $\Delta T$\\ \hline
1 & 45, 0 & 400 & 0, 0, 35 & \textbf{4.4} & \textbf{0.4} & \textbf{0.79} & \textbf{0.8} & 15.0 & 14.8 & 2.04 & 3.35 \\ \hline
2  & 45, 5 & 400 & 2, 3, 35 & \textbf{5.70} & \textbf{3.10} & \textbf{0.97} & \textbf{0.86}  & 22.1 & 7.90 & 2.87 & 2.08\\ \hline
3 & \multicolumn{3}{c||}{Set 2 \textit{w/} 4 bad poses}& \textbf{3.44} & \textbf{5.70} & \textbf{1.14} & \textbf{3.26}  & 26.60 & 16.70 & 3.63 & 3.96 \\ \hline
4 & \multicolumn{3}{c||}{Set 3 \textit{w/o} bad poses}& \textbf{3.20} & \textbf{5.50} & \textbf{1.05} & \textbf{0.96}  & N/A & N/A & N/A & N/A\\ \hline
\end{tabular}}
\begin{tablenotes} \item [*] Ground truth principal point: $(320, 240).$\end{tablenotes}
\end{threeparttable}
\end{minipage}
\vspace{-5pt}
\end{table*}

\subsubsection{Synthetic Data without Noise}
Under the setting without noise, Figure~\ref{fig:8_planes_synthetic}(b) shows the resultant eight (perfect) principal lines obtained in closed-form using (\ref{eq:pp_line}), which intersect exactly at the principal point. One can see that due to the symmetry in their orientations, only four (pairs of) principal lines can be seen from the illustration.
Perfect estimation results for focal length (FL) and other extrinsic parameters are obtained for both the proposed and Zhang's method for this simple setting. Note that real numbers are used to represent all numerical values during the associated computations, while the input images and the final results shown in Figure \ref{fig:8_planes_synthetic} are illustrated with virtual 640$\times$480 images.

\subsubsection{Synthetic Data with Noise and Variation of Poses}\label{sec:syn_noise_pose}

To investigate the robustness of the proposed approach, noises are added to the point features used to find $\mathbf{H}$ in (\ref{eq:homography}), which are the only source of interference that may affect the correctness of each principal line. Figure \ref{fig:8_planes_synthetic}(c) illustrate calibration results similar to those shown in Figure \ref{fig:8_planes_synthetic}(b), with noises uniformly distributed between $\pm1.0$ pixels added to $x$ and $y$ coordinates of the corners shown in Figure \ref{fig:8_planes_synthetic}(a) to simulate point feature localization errors resulted from image digitization.

For more systematic error analysis, and also taking into account the influence from different poses of the calibration plane, similar simulations are performed for two different noise levels with $\gamma$ ranging from $\ang{5}$ to $\ang{85}$ (with $\Delta\gamma=\ang{5}$ and $\Delta\alpha=45\degree$), and repeated 20 times for each pose of the calibration plane
. Because of the simplicity of the proposed approach, a total of 17×20 = 340 principal points are estimated in 4.5 seconds for each noise level, including the analysis of $340\times8=2720$ images to generate the same number of principal lines.  

Figures~\ref{fig:pp_err}(a) and (b) show means and standard deviations of estimation error (in image pixels), respectively, for the above simulation. It readily observable that larger noises will result in less accurate calibration results which are also less robust. 
Aside from the statistical comparison, we found that better results are generated for poses of calibration plane away from 2 degenerated conditions in Figure \ref{fig:wcs_ipcs}, i.e., $\Pi_1 \parallel \Pi_2$ and $\Pi_1 \perp \Pi_2$. Moreover, according to the results in Figure~\ref{fig:pp_err}, it is reasonable to suggest that: (i) the best values of $\gamma$ are around $\ang{45}$. Based on such observation, $\gamma$ is selected to be approximately $\ang{45}$ for the following experiments. Furthermore, as the principal point is derived via least square solution (\ref{eq:pp_sol}) from all the principal lines, it is also suggested that: (ii) it is better to distribute $\alpha$ uniformly
between $\ang{0}$ and $\ang{180}$.

\subsubsection{Full Camera Calibration for Fixed Focal Length}\label{sec:real_fix_focal}
As for a more complete error analysis for the estimation of all parameters of a camera with fixed focal length, more general \textbf{IPCS-WCS} configurations (also with $\Delta \alpha = \ang{45}$) are considered, as listed in Table \ref{tab:synthetic_data_ours_vs_zhang} for the first two datasets, with additive noises of $\pm1.0$ pixels. One can see that smaller estimation errors are achieved for all parameters by the proposed approach, possibly due to the simplicity of the formulations established in Section \ref{sec:para_derive}, compared with those given in \cite{Zhang2000}. (Note that $\Delta FL$ is computed for the average value of FLs each obtained for a single input image for the  proposed approach, and for the mean value of $f_x$ and $f_y$ for Zhang's method.)

As for the robustness of camera calibration, it is possible for the proposed geometric-based approach to improve the parameter estimation by screening out calibration pattern with bad poses. For example, the 3\textsuperscript{rd} dataset in Table \ref{tab:synthetic_data_ours_vs_zhang} is obtained by replacing four input patterns of the 2\textsuperscript{nd} dataset with four unfavorable ones ($\gamma < \ang{20}$), resulting in less accurate estimates for most parameters. Nonetheless, by removing such patterns using (\ref{eq:gamma_prime}), as shown in Table \ref{tab:synthetic_data_ours_vs_zhang} with the last (4\textsuperscript{th}) dataset, the calibration can be improved for all parameters.
\begin{table*}[!htb]
\small \addtolength{\tabcolsep}{-4.4pt}
\centering
\captionof{table}{Comparison of full camera calibration (varied focal length)}
\label{tab:synthetic_data_ours_vs_zhang_different_fl}
\begin{threeparttable}
\resizebox{0.99\linewidth}{!}{%
\begin{tabular}{|c|c|c|c|c||c|c|c|c||c|c|c|c|}
\hline
\multirow{2}{*}{Set} & \multirow{2}{*}{Noise} & \multicolumn{3}{c|}{Given params.} & \multicolumn{4}{c|}{Ours} & \multicolumn{4}{c|}{Zhang's} \\ \cline{3-13} 
 & & $\gamma$, $\beta$ & FL & $t_x$, $t_y$, $t_z$ & $\Delta PP$ & $FL_{est}$ (for 8 calibration patterns)  & $\Delta R$ & $\Delta T$  & $\Delta PP$ & $FL_{est}$ & $\Delta R$ & $\Delta T$\\ \hline
5 & $\times$ & 40, 10 & 400 \& 440 & 0, 0, 35 & \textbf{0.0} & \textbf{400.0}, \textbf{400.0}, \textbf{400.0}, \textbf{400.0}, \textbf{440.0}, \textbf{440.0}, \textbf{440.0}, \textbf{440.0} & \textbf{0.0}&\textbf{0.0}&4.9 & 320.4&1.61&3.85\\ \hline
6 & \checkmark & 40, 10 & 400 \& 440 & 0, 0, 35 & \textbf{5.2} & 402.5, 383.4, 395.1, 409.2, \textbf{430.7}, \textbf{449.1}, \textbf{422.9}, \textbf{442.3} & \textbf{0.89} & \textbf{0.84} & 14.0 & 399.4&1.63&3.02\\ \hline
\end{tabular}}
\end{threeparttable}
\vspace{-10pt}
\end{table*}

\subsubsection{Full Camera Calibration for Varied Focal Length}\label{sec:syn_focal}
As one of the key feature (\textbf{F4}) mentioned in Section \ref{sec:intro}, the proposed method can calibrate cameras with non-fixed focal length (FL), while Zhang's method is not designed to cope with such situation. Table \ref{tab:synthetic_data_ours_vs_zhang_different_fl} shows the calibration results obtained with the proposed approach, as well as those from Zhang's method. It is readily observable that even under the noise-free condition, significant estimation errors can already be observed in the latter, although perfect estimations are achieved with the proposed approach. As for the noisy case,  while our results will have some expected estimation errors, Zhang's method may generate extraneous errors for some, if not all, parameters. In either case, the major difference of calibration performance is in the estimation of focal length, which is greatly constrained by the ability to cope with varied focal length during the image acquisition process. 

\subsection{Performance Evaluation Using Real Data}
\begin{figure}
  \begin{minipage}{0.5\linewidth}
    \begin{tabular}{cc}
    	\centering
    	\includegraphics[width=0.4\linewidth]{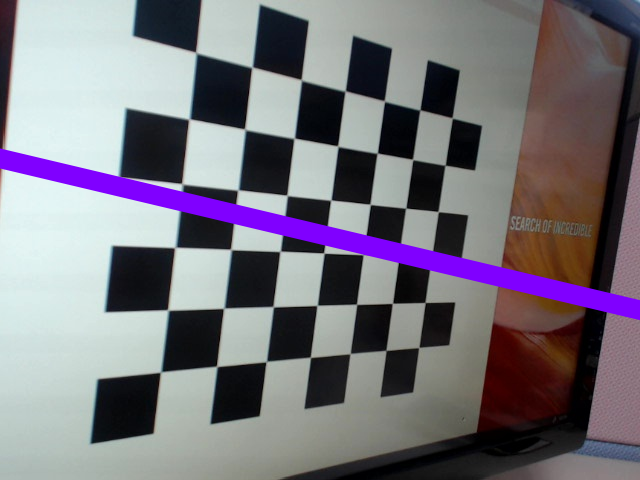}
        & \includegraphics[width=0.4\linewidth]{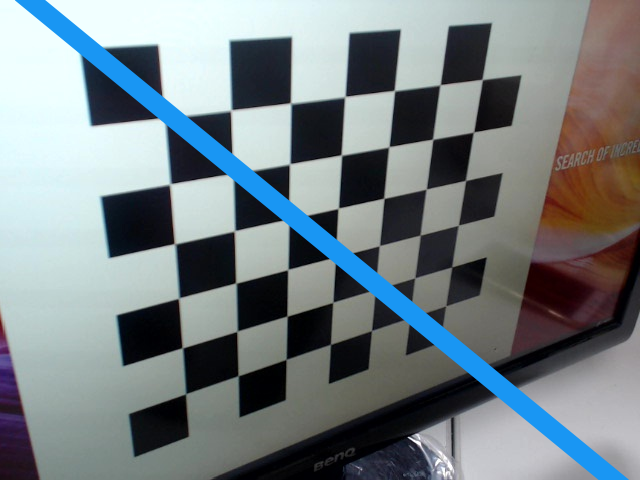}\\
        \includegraphics[width=0.4\linewidth]{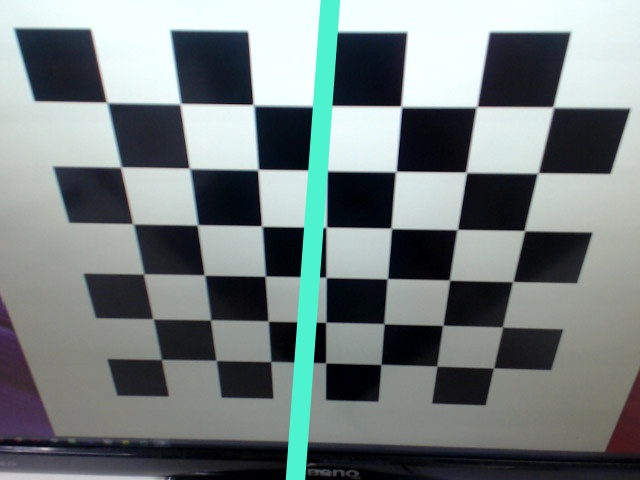}
        & \includegraphics[width=0.4\linewidth]{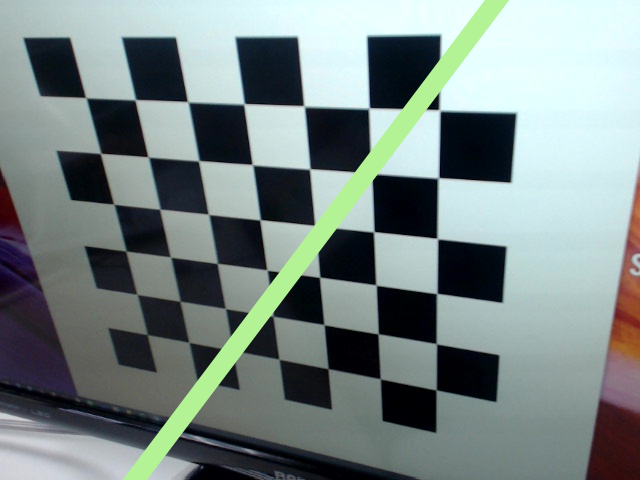}
    \end{tabular}\\
    \centering
    \scriptsize{(a)}
  \end{minipage}%
   \begin{minipage}{0.5\linewidth}
    \begin{tabular}{c}
    \centering
  		\includegraphics[width=0.8\linewidth]{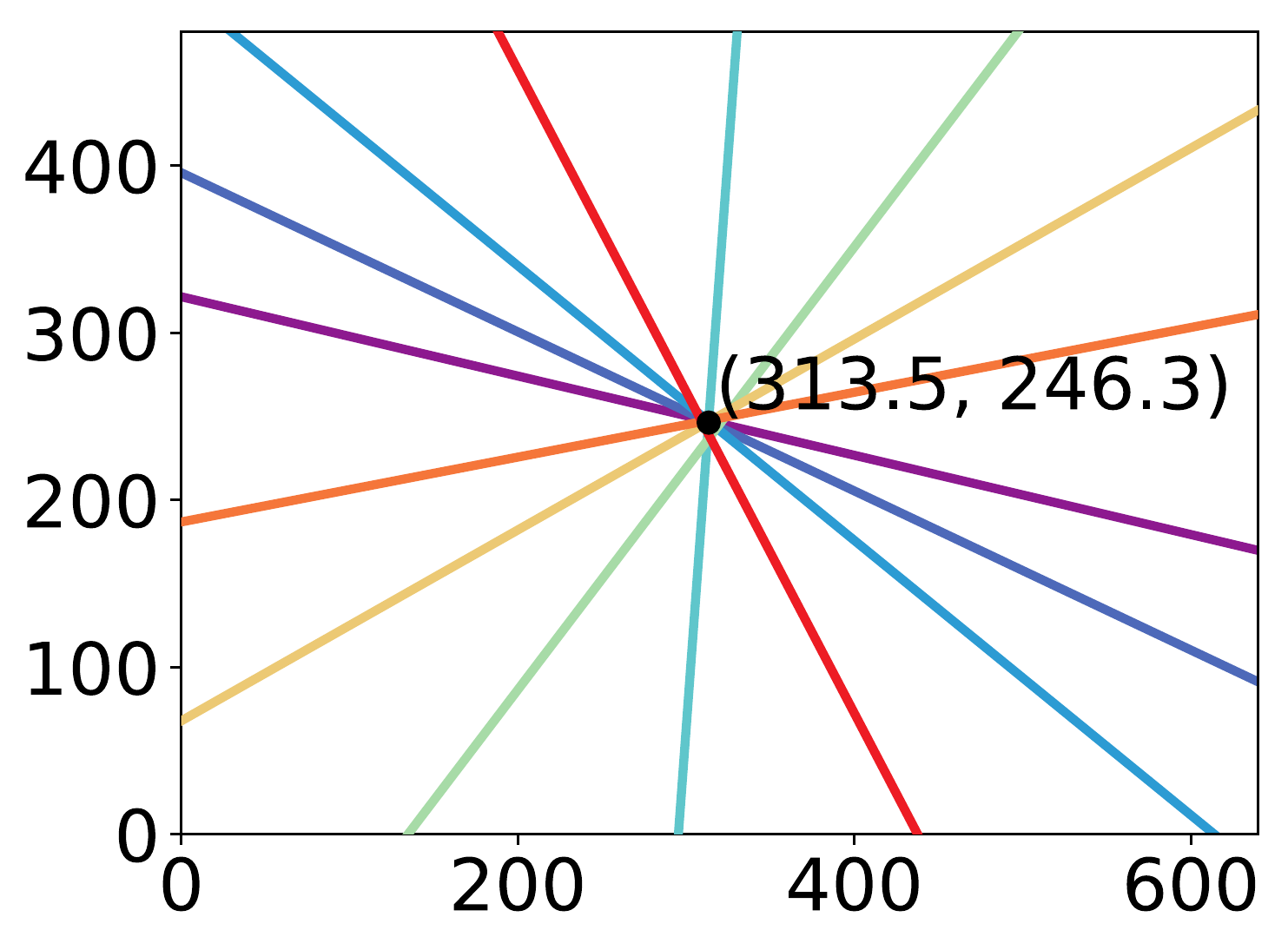}
    \end{tabular}\\
    \centering
    \scriptsize{(b)} 
  \end{minipage}
 
  \caption{(a) Four samples of eight good calibration images with $\alpha$ nearly uniformly distributed from $\ang{0}$ - $\ang{180}$. (b) Eight principal lines, each obtained from an calibration images.}
  \label{fig:good_cali_img_set}
\end{figure}
For performance evaluation of the proposed calibration method under realistic conditions, 
7$\times$8 checkerboard images are employed as calibration patterns. Similar to the experiments
considered in Section \ref{sec:3_1}, we demonstrate the flexibility of the proposed work over Zhang's method by comparing the respective performance under 2 different experiment setups, i.e., (a) using fixed focal length and (b) using different focal lengths throughout the image acquisition process. For setup (a), another approach of outlier removal, which is based on the RMSE of the estimation of principal point in (\ref{eq:pp_sol}), will be demonstrated to filter out ill-posed calibration planes to further improve the stability and robustness of the result.  

\subsubsection{Real Data with Fixed Focal Length}
For camera calibration considered in this section, the focal length of the camera is fixed while calibration patterns are captured. As suggested in Section \ref{sec:syn_noise_pose}, a good set of images should have $\gamma \cong \ang{45}$ and $\alpha$ should be nearly uniformly distributed between $\ang{0}$ and $\ang{180}$. Figure~\ref{fig:good_cali_img_set}(a) shows four of eight images thus obtained with a Logitech Webcam camera with a image resolution of 640$\times$480, while Figure~\ref{fig:good_cali_img_set}(b) shows
a total of 8 principal lines, with (313.5, 246.3) being estimated as the location of the principal point. 
The estimated principal point, which is near the center of the image, along with the estimated focal length (FL) are shown as the 1\textsuperscript{st} set (set 7) of data of Table \ref{table:real_fix_focal}. Under this near ideal circumstances, both our method and Zhang's method produce similar results. Note that rotation and translation are not reported due to limited space and can be found in supplementary materials.

\begin{figure}
\centering
    \begin{tabular}{cc}
        \includegraphics[width=0.36\linewidth]{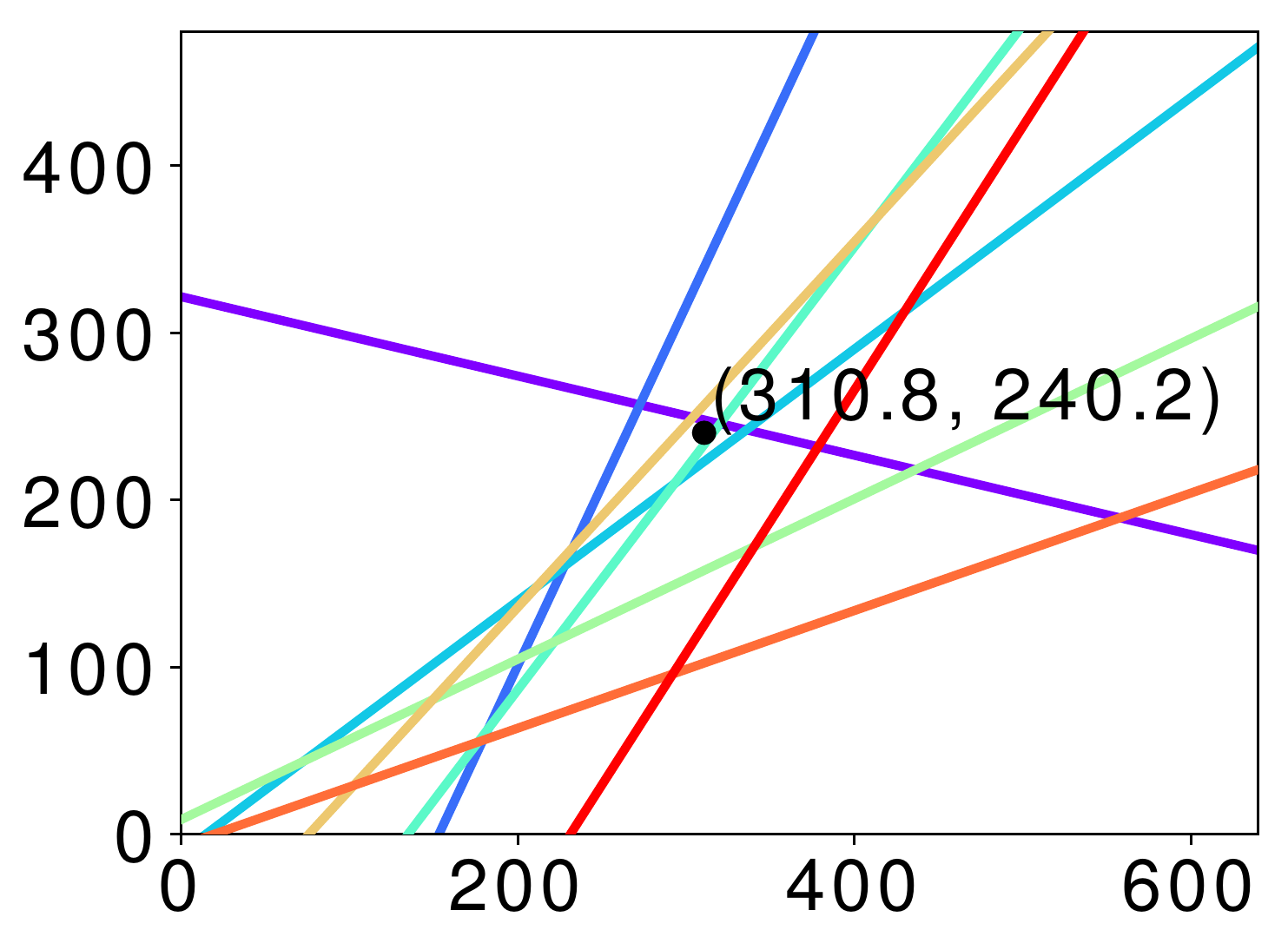} & 
        \includegraphics[width=0.36\linewidth]{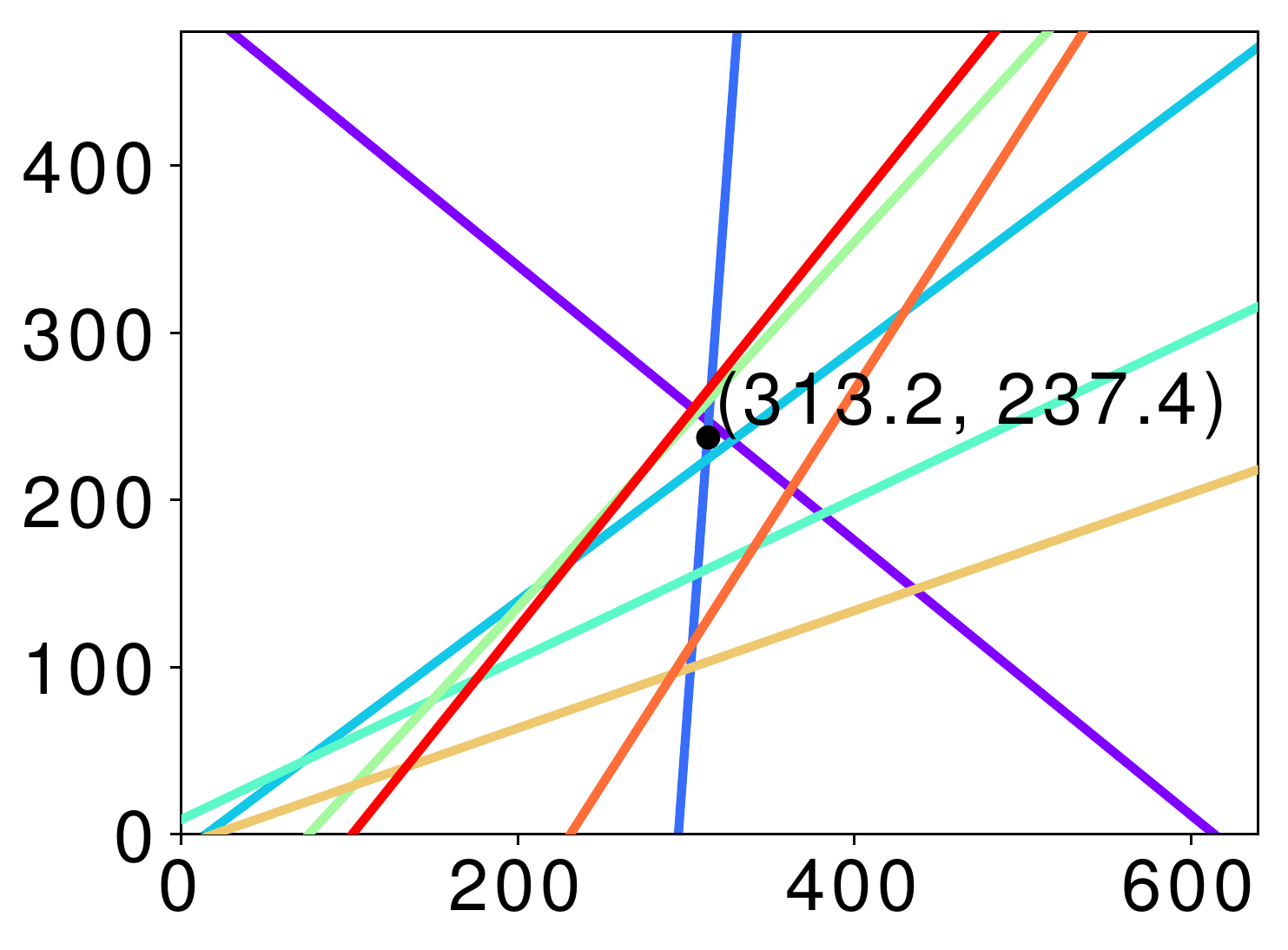}\\
        \scriptsize{(a)} & \scriptsize{(b)} \\
        \includegraphics[width=0.36\linewidth]{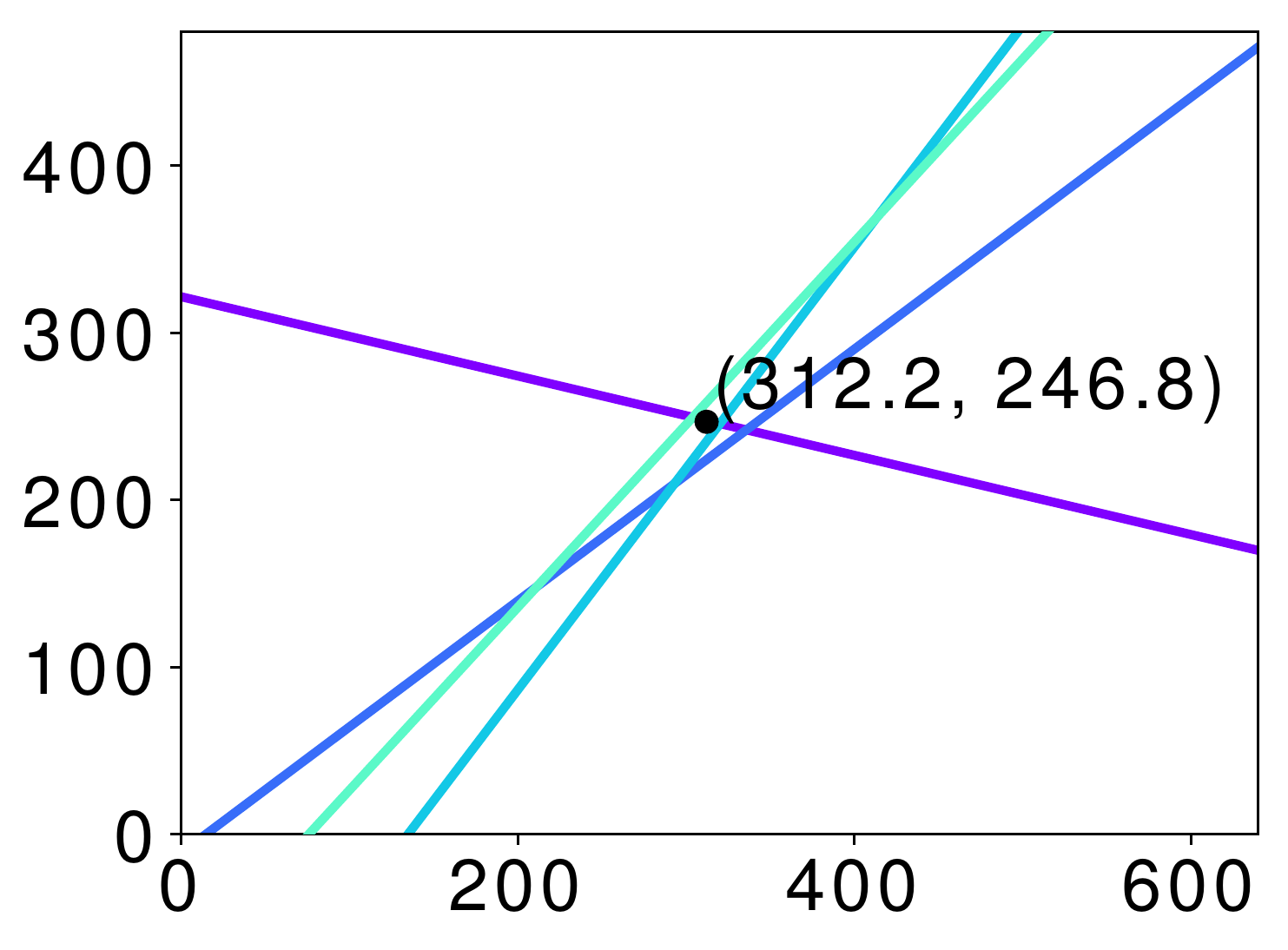} & 
        \includegraphics[width=0.36\linewidth]{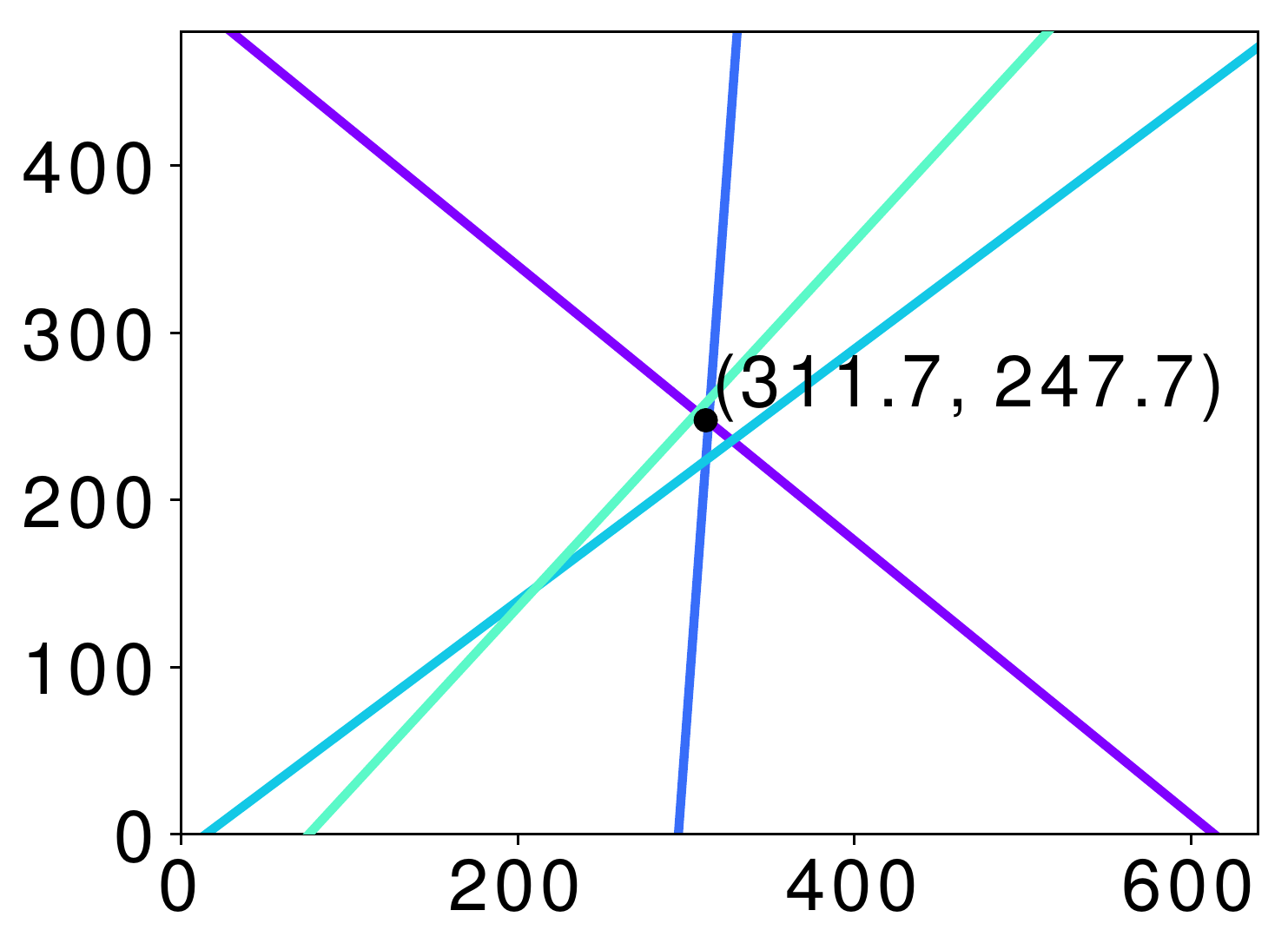} \\
        \scriptsize{(c)} & \scriptsize{(d)} 
    \end{tabular}
    \caption{Selected sets of principal lines due to ill-posed images in (a) Set 8, (b) Set 9, and their counterparts after performing outlier removal: (c) Set 11 and (d) Set 12.}
    \label{fig:real_with_bad_poses}
 
\end{figure}

To evaluate the sensitivity of calibration to unfavorable (ill-posed) calibration patterns, results of three more datasets (Sets 8-10) are also included in Table \ref{table:real_fix_focal}, each obtained by replacing some good patterns in Set 7 with unfavorable ones. The adverse influence of such replacements are readily observed from the dramatic increments of RMSE/STD of the proposed approach, e.g., for principal lines shown in Figures \ref{fig:real_with_bad_poses}(a) and (b), and the estimation errors in PP/FL of Zhang's method. Nonetheless, the proposed approach seems to be more robust as the average values of PP and FL are not affected as much.

\begin{table}
\centering
\small \addtolength{\tabcolsep}{-4.4pt}
\caption{
Comparison of camera calibration (real data, fixed focal length)
}
\begin{threeparttable}
\resizebox{0.99\hsize}{!}{
\begin{tabular}{|c|c|c||c|c|}
\hline
\multirow{2}{*}{Set}  & \multicolumn{2}{c||}{Ours} & \multicolumn{2}{c|}{Zhang's} \\ \cline{2-5} 
 &  PP/RMSE  & $FL_{avg}$/STD  & PP     & FL   \\ \hline
7  & (313.5, 246.3)/2.77 & 617.1/7.8 & (317.6, 248.9) & 616.8 \\ \hline
8  & (\textbf{310.8}, \textbf{240.2})/59.8 & \textbf{610.2}/29.7 & (342.2, 235.8) & 631.8 \\ \hline
9  & (\textbf{313.2}, \textbf{237.4})/56.4 & \textbf{611.5}/26.3 & (350.1, 230.9) & 634.8 \\ \hline
10 & (\textbf{307.1}, \textbf{235.7})/30.7 & \textbf{619.6}/30.7 & (335.1, 237.0) & 628.2\\ \hline
\end{tabular}}
\end{threeparttable}
\label{table:real_fix_focal}
\vspace{-10pt}
\end{table}

\begin{table}
\centering
\small \addtolength{\tabcolsep}{-4.4pt}
\caption{Calibration results improved from Sets 8-10 in Table \ref{table:real_fix_focal} via outliers removal.
}
\begin{threeparttable}
\begin{tabular}{|c|c|c|}
\hline
\multirow{2}{*}{Set}  & \multicolumn{2}{c|}{Ours} \\ \cline{2-3} 
 &  PP/RMSE    & $FL_{avg}$/STD\\ \hline
11  & (312.2, 246.8)/10.5 & 608.5/23.3\\ \hline 
12  & (311.7, 247.7)/10.2 & 607.2/20.9 \\ \hline
13 & (310.1, 244.6)/8.5 & 614.2/22.5 \\ \hline
\end{tabular}
\end{threeparttable}
\label{table:real_fix_focal_after_outliers_removal}
\end{table}

\begin{table*}
\small \addtolength{\tabcolsep}{-4.4pt}
\centering
\caption{
Comparison of camera calibration (real data, varied focal length)
}
\begin{tabular}{|c|c|c|c||c|c|}
\hline
\multirow{2}{*}{Set} & FL & \multicolumn{2}{c||}{Ours} & \multicolumn{2}{c|}{Zhang's} \\ \cline{3-6} 
  & (mm)  & PP  & FL & PP  & FL \\ \hline
14 & 39 & (1917.6, 1270.5) & 3851.7, 3787.6, 3817.9, 3836.3, 3837.9, 3835.6  & (1923.0, 1274.5) & 3822.3 \\ \hline
15  & 50 & (1920.6, 1263.8) &  4770.0, 4741.9, 4727.2, 4669.1, 4721.7, 4707.1  & (1919.1, 1275.6) & 4712.5\\ \hline
16  & Mixed & (\textbf{1916.8}, \textbf{1266.0}) & 3791.5, 3823.2, 3841.7, 4734.7, 4710.2, 4766.1  & (1897.0, 1416.8) & 3953.0\\ \hline
\end{tabular}\label{table:real_nonfix_focal}
\end{table*}

On the other hand, it is possible for the proposed approach to remove the above problematic calibration patterns, similar to that performed in Sec. \ref{sec:real_fix_focal} for synthetic data. Specifically, Sets 11, 12 and 13 in Table \ref{table:real_fix_focal_after_outliers_removal} are obtained by simply screening out possible outliers in Sets 8, 9 and 10, respectively, whose RMSE are greater than 15. One can see that most estimations are improved, and with both RMSE/STD reduced. Figures \ref{fig:real_with_bad_poses}(c) and (d) show such outlier removal results for their counterparts shown in Figures \ref{fig:real_with_bad_poses}(a) and (b), respectively.

\begin{figure}
  \begin{minipage}{0.49\linewidth}
    \begin{tabular}{cc}
    	\centering
    	\includegraphics[width=0.36\linewidth]{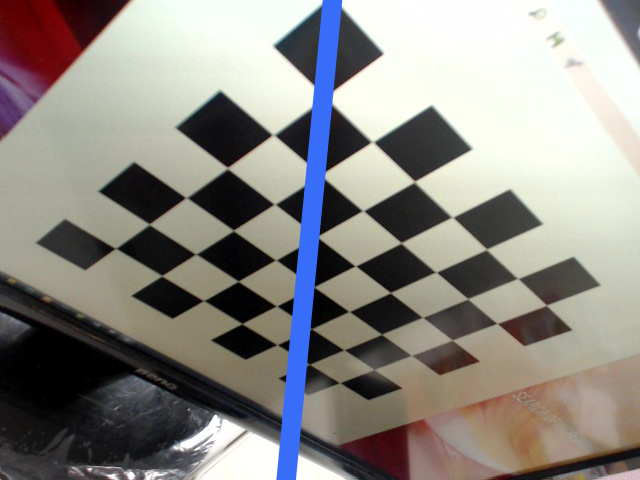}
        & \includegraphics[width=0.36\linewidth]{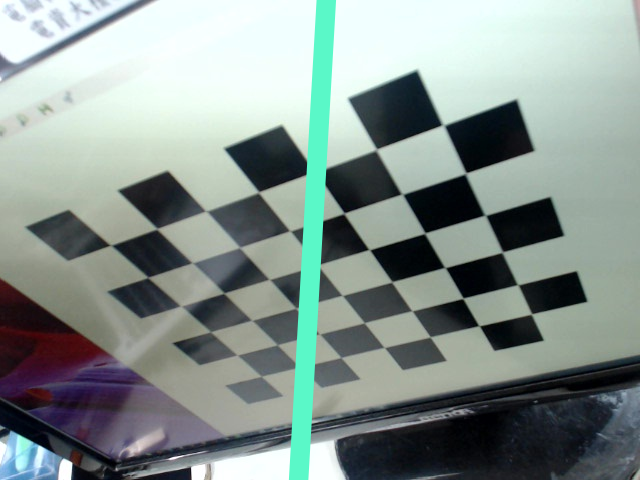}\\
        \includegraphics[width=0.36\linewidth]{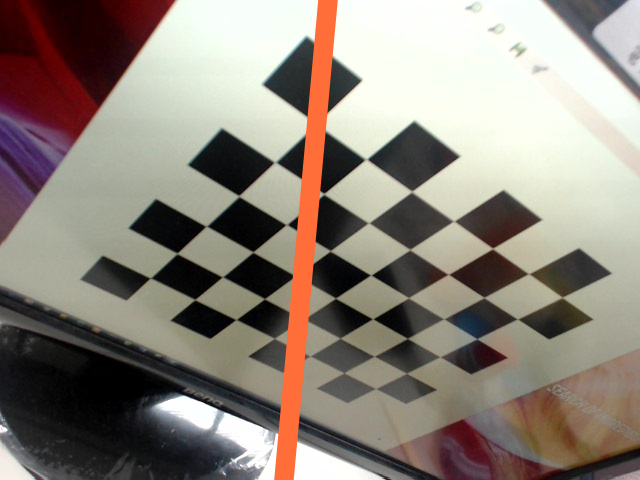}
        & \includegraphics[width=0.36\linewidth]{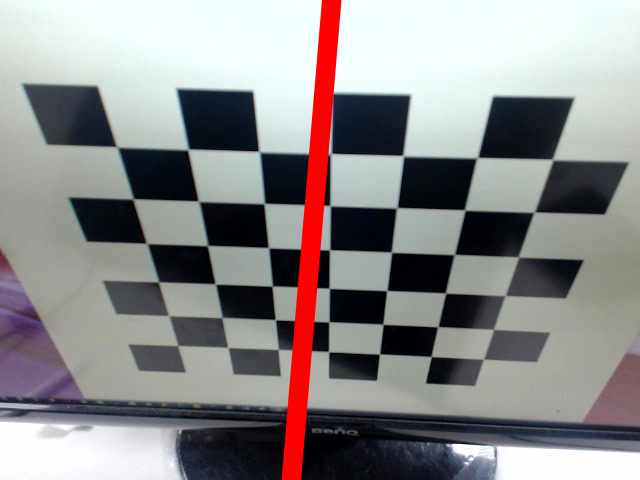}
    \end{tabular}\\
    \centering
    \scriptsize{(a)}
  \end{minipage}%
   \begin{minipage}{0.49\linewidth}
    \begin{tabular}{c}
    \centering
  		\includegraphics[width=0.85\linewidth]{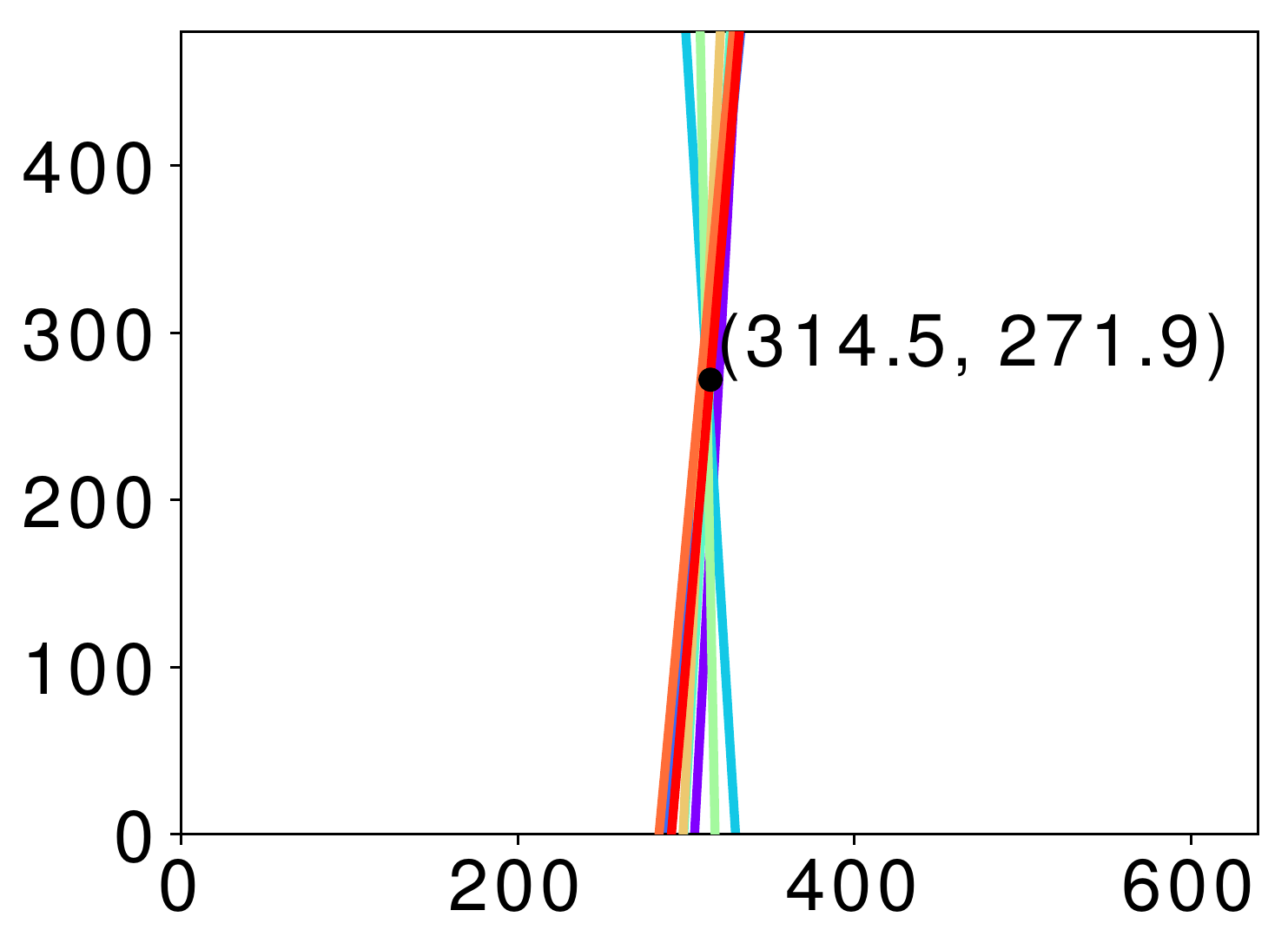}
    \end{tabular}\\
    \centering
    \scriptsize{(b)} 
  \end{minipage}
 
  \caption{(a) Four samples of eight good calibration patterns with good poses  (b) the eight nearly parallel principal lines.}
  \label{fig:good_img_bad_set}
\end{figure}

Beside ill-posed calibration patterns, a set of good patterns which violates guideline (ii) mentioned in Sec. \ref{sec:syn_noise_pose} may also gives unreliable estimation results, as shown in Figures \ref{fig:good_img_bad_set}(a) and (b) for the calibration patterns and the corresponding principal lines, respectively. However, such condition can be easily detected by using azimuth angle of the principal lines in (\ref{eq:rotate_angle}), and new calibration images can be retaken to generate more reliable calibration results.

\subsubsection{
Camera Calibration with Varied Focal Length
}
In this subsection, calibration of cameras with varied focal length (FL) is considered, which is a more challenging but corresponds to fairly common situation in real world scenarios. Table \ref{table:real_nonfix_focal} shows the calibration results for three sets of calibration patterns which are captured by a Canon 5D camera with an image resolution of 3840$\times$2560.\footnote{Due to limited space, images of calibration patterns similar to those shown in Figure \ref{fig:good_cali_img_set} are provided in the supplementary material.} As guidelines mentioned in Sec. \ref{sec:syn_noise_pose} for selecting good poses of the calibration pattern are closely followed, satisfactory calibration results are obtained with both methods for the first two datasets (Sets 14 and 15), each established for a fixed (but different) FL.


On the other hand, consider the last dataset (Set 16) in Table \ref{table:real_nonfix_focal}, which corresponds to a mixture of calibration patterns from Set 14 and Set 15, with half of them obtained from the 1\textsuperscript{st} half of the former and the other half from the 2\textsuperscript{nd} half of the latter. It is readily observable that our approach still performs satisfactorily and generates results similar to those for Sets 14 and 15. However, unacceptable results are obtained with Zhang's method which assumed fixed FL. In particular, the estimated FL (3953.0) is quite different from one of the two FLs obtained for the fixed cases (3822.3) and \textbf{\textit{very}} different from the other one (4712.5). Moreover, the estimation of principal point is also impaired quite seriously under such situation, i.e., with a deviation of more than 20 pixels (140 pixels) in the horizontal (vertical) directions. (Similar problems can be expected for the estimation of extrinsic parameters as well.)

\vspace{-1mm}
\section{Conclusion}\label{sec:conclusion}
In this paper, we have made a major attempt to establish a new camera calibration procedure based on a geometric perspective. The proposed approach resolves two main issues associated with the widely used Zhang's method, which include the lack of clear hints of appropriate pattern poses and the limitation of its applicability imposed by the assumption of fixed focal length. The main contribution of this work is to provide a closed-form solution to the calibration of extrinsic and intrinsic parameters based on the analytically tractable principal lines, with intersection of such lines being the principal point while each of them conveniently representing the relative 3D orientation and position (up to one degree of freedom for both) between the image plane and a calibration plane for a corresponding \textbf{IPCS-WCS} pair. Consequently, computations associated with the calibration can be greatly simplified, while useful guidelines to avoid outliers in the computation can be established intuitively. Experimental results for both synthetic and real data clearly validate the correctness and robustness of the proposed approach, with both compared favorably with Zhang’s method, especially in terms of the possibilities to screen out problematic calibration patterns as well as the ability to cope with the situation of varied focal length. More applications of this new technique of camera calibration are currently under investigation.

\clearpage
{\small
\bibliographystyle{ieee}
\bibliography{main}
}

\end{document}